\documentclass[lettersize,journal]{IEEEtran}
\usepackage{amsmath,amsfonts}
\usepackage{algorithmic}
\usepackage{algorithm}
\usepackage{array}
\usepackage[caption=false,font=normalsize,labelfont=sf,textfont=sf]{subfig}
\usepackage{textcomp}
\usepackage{stfloats}
\usepackage{url}
\usepackage{verbatim}
\usepackage{graphicx}

\usepackage{microtype}

\usepackage{tikz}
\usetikzlibrary{arrows.meta,positioning,fit}

\usepackage[capitalize]{cleveref}
\usepackage[style=ieee,doi=false,isbn=false,date=year,backend=biber,maxbibnames=15,maxcitenames=2,mincitenames=1,uniquelist=false,uniquename=false,giveninits=true,useprefix=true]{biblatex}

\usepackage{makecell}
\usepackage{tabularx}
\usepackage{multirow}
\usepackage{array}
\usepackage{multicol}
\usepackage{adjustbox}
\usepackage{makecell}
\usepackage{booktabs}
\usepackage{rotating}
\usepackage{xcolor}
\usepackage[nolist,nohyperlinks]{acronym}
\usepackage[per-mode=symbol]{siunitx}
\usepackage{tikz}
\usepackage[colorinlistoftodos]{todonotes}
\usepackage{mdframed}

\usepackage{pifont} 
\usepackage{wasysym} 
\usepackage{amssymb} 
\newcommand{\fullcircle}{\CIRCLE} 
\newcommand{\halfcircle}{\LEFTcircle} 
\newcommand{\emptycircle}{\Circle} 

\newcommand{\cstart}{\color{black}}

\begin{acronym}[AAAAAAAA]
    \acro{av}[AV]{automated vehicle}\acroindefinite{av}{an}{an}
    \acro{ad}[AD]{automated driving}\acroindefinite{ad}{an}{an}
    \acro{btn}[BTN]{brake threat number}
    \acro{CAE}{convolutional autoencoders}
    \acro{dhd}[DHD]{directed hausdorff distance}
    \acro{HARA}{hazard analysis and risk assessment}
    \acro{id}[ID]{in-distribution}\acroindefinite{id}{an}{an}
    \acro{KDE}{kernel density estimation}
    \acro{nn}[NN]{neural network}\acroindefinite{nn}{an}{an}
    \acro{odd}[ODD]{operational design domain}\acroindefinite{odd}{an}{an}
    \acro{OEM}{original equipment manufacturer}\acroindefinite{oem}{an}{an}
    \acro{ood}[OoD]{out-of-distribution}\acroindefinite{ood}{an}{an}
    \acro{pca}[PCA]{principal component analysis}
    \acro{pdf}[PDF]{probability density function}
    \acro{SCONE}{semantic and covariate OoD learning via energy margins}
    \acro{shnncad}[SHNN-CAD]{sequential Hausdorff nearest-neighbor conformal anomaly detector}
    \acro{STPA}{system theoretic process analysis}
    \acro{SVDD}{support vector data description}
    \acro{ttc}[TTC]{time-to-collision}
    \acro{VAE}{variational autoencoders}
    \acro{auroc}[AURCO]{area under the ROC curve}
    \acro{ap}[AP]{average precision}
    \acro{tpr}[TPR]{true positive rate}
    \acro{fpr}[FPR]{false positive rate}
    \acro{fnr}[FNR]{false negative rate}
    \acro{auc-prc}[AUC-PRC]{area under the precision-recall curve}
    \acro{auc-roc}[AUC-ROC]{area under the curve of the receiver operating characteristics}
\end{acronym}

\begin{document}

\title{Edge Case Detection in Automated Driving:\\ Methods, Challenges and Future Directions}

\author{Saeed Rahmani,~\IEEEmembership{Graduate Student Member,~IEEE,} Sabine Rieder, Erwin de Gelder, Marcel Sonntag, Jorge Lorente Mallada, Sytze Kalisvaart, Vahid Hashemi, Bart van Arem, Simeon C. Calvert
\thanks{Saeed Rahmani, Simeon C. Calvert, and Bart van Arem are with the Department of Transport \& Planning, Delft University of Technology, Delft, The Netherlands (emails: s.rahmani@tudelft.nl; s.c.calvert@tudelft.nl; b.vanarem@tudelft.nl). Sabine Rieder is with the School of Computation, Information and Technology at Technical University of Munich, Munich, Germany, and Faculty of Informatics at Masaryk University, Brno, Czechia (email: sabine.rieder@tum.de). Erwin de Gelder and Sytze Kalisvaart are with the Department of Integrated Vehicle Safety at Netherlands Organization for Applied Scientific Research (TNO), Helmond, The Netherlands (emails: erwin.degelder@tno.nl; sytze.kalisvaart@tno.nl). Marcel Sonntag is with the Institute for Automotive Engineering (ika), Department of Vehicle Intelligence \& Automated Driving at RWTH Aachen University, Aachen, Germany (email: marcel.sonntag@ika.rwth-aachen.de). Jorge Lorente Mallada is with the Technical Affairs Planning \& Safety Research team at Toyota Motor Europe, Brussels, Belgium (email: jorge.lorente.mallada@toyota-europe.com), and Vahid Hashemi is with Audi AG, Ingolstadt, Germany (email: vahid.hashemi@audi.de).
\\
Digital Object Identifier (DOI): 10.1109/TITS.2026.3715674}}

\markboth{Accepted for publication at IEEE Transactions on Intelligent Transportation Systems. DOI: 10.1109/TITS.2026.3715674}%
{Rahmani \MakeLowercase{\textit{et al.}}: A Systematic Review of Edge Case Detection in Automated Driving: Methods, Challenges and Future Directions}


\maketitle

\begin{abstract}
\Acp{av} promise to enhance transportation safety and efficiency. However, ensuring their reliability in real-world conditions remains challenging, particularly due to rare and unexpected situations known as edge cases. While numerous approaches exist for detecting edge cases, a comprehensive survey reviewing these techniques is lacking. This paper bridges this gap by presenting a hierarchical review and systematic classification of edge case detection and assessment methodologies. Our classification is structured on two levels: first, by AV modules, including perception and trajectory-related (encompassing prediction, planning, and control) subsystems; and second, by underlying methodologies and theories guiding these techniques. Furthermore, we introduce ``knowledge-driven'' approaches, which complement data-driven methods by leveraging expert insights and domain knowledge to identify cases absent in training datasets. We then examine techniques and metrics for evaluating edge case detection methods, including detection performance, practical deployment (e.g., computational overhead), and domain-specific measures (e.g., crash rates and severity analysis). We conclude by highlighting key challenges for edge case detection, including data availability and quality issues, validation and interpretability limitations, the simulation-to-real gap, and computational constraints. The hierarchical classification and review of methods and assessment techniques in this survey enable modular and targeted testing frameworks by guiding the selection of detection methods for specific \ac{av} subsystems while considering methodological principles. It also supports practical testing by facilitating scenario generation in simulation and focused subsystem validation in the real world.
\end{abstract}


\acresetall  
\section{Introduction}
\Acp{av} have seen significant progress towards deployment readiness. However, key challenges, particularly those associated with their rigorous testing and validation, have hindered their widespread adoption. Validating \ac{av} systems requires extensive testing across diverse conditions and addressing ethical and legal accident liabilities \cite{ santoni2021four}. {\cstart Given the very large number of scenarios that \acp{av} may encounter, it is impractical to conduct physical tests for every possible situation. This limitation highlights the critical role of ``edge cases'' in \ac{av} development. Edge cases are novel or rare situations that occur frequently enough in full-scale deployments to require specific design attention so they can be handled reasonably and safely by automated driving functions~\cite{Vater2023ADriving} (cf. \cref{sec:definition} for formal definition)}. Examples include extreme weather, unexpected object behaviours, and unknown objects.

Edge cases are critical for the development and testing of \acp{av}. Because AV training datasets are dominated by common driving situations, systems are inherently optimised for routine conditions \cite{zhang2025review}; edge cases, which lie near the limits of system performance, are therefore under-represented yet disproportionately consequential. Systematically identifying and addressing them enables targeted performance improvements spanning safety, passenger comfort, and operational efficiency. This need is also reflected in regulatory frameworks: ISO 21448 (SOTIF) \cite{iso21448} specifically addresses the reduction of unknown hazardous scenarios by identifying ``triggering conditions,'' of which edge cases are prime examples. Furthermore, public acceptance of \acp{av} depends on demonstrated reliability in unforeseen circumstances. Unresolved edge cases manifest as false positive detections, which trigger unnecessary interventions that compromise comfort and traffic flow, or false negative detections that cause abrupt reactions and safety-critical failures.

While numerous identification methods exist, the field lacks a holistic review and synthesis of these methods, their strengths, limitations, and cross-subsystem applicability. Moreover, practical deployment challenges, such as sim-to-real gaps, data availability constraints, and scalability issues, remain inadequately addressed in existing literature. This gap hinders AV developers and researchers from understanding the trade-offs between different approaches and selecting appropriate methods for their specific applications. 

This paper aims to bridge this gap by systematically classifying edge case detection and assessment methods, identifying their respective strengths and weaknesses, and discussing challenges and future research directions. To achieve this, we adopt a hierarchical approach. First, we classify edge case detection methods based on the corresponding modules of the \ac{av} stack. We distinguish between ``perception-related'' edge cases, which arise from challenges in sensing and interpreting the driving environment, and ``trajectory-related'' edge cases, which stem from motion prediction, planning, and control processes within the \ac{av}. Next, we break down the classification of methods based on their underlying principles for detecting edge cases. Beyond that, we introduce a largely overlooked class of edge case detection approaches, namely ``knowledge-driven,'' which leverages expert insights to detect or generate new edge cases. Knowledge-driven methods enable proactive identification of edge cases that may never appear in crash databases or training datasets. Beyond this overview of detection methods, we review evaluation metrics and assessment techniques for measuring edge case detection performance, and discuss the key challenges and future research directions for advancing edge case detection in the \ac{av} domain.

By adopting this multi-faceted approach, our survey makes several key contributions to the field: (1) a systematic taxonomy of edge case detection methods across both perception-related and trajectory-related edge cases, which builds upon prior surveys that emphasise perception anomalies  \cite{Breitenstein2020SystematizationDriving, Heidecker2021AnDrivingb} or safety-critical scenario identification \cite{Bogdoll2022AnomalySurvey, Zhang2022Finding}, (2) the introduction of knowledge-driven approaches as a complement for data-driven methods, (3) an overview of assessment techniques and metrics across different edge case categories, and (4) a holistic analysis of challenges and future directions that encompass all AV subsystems and edge case types.

We believe this survey contributes to the \ac{ad} community in several ways. {\cstart Firstly, its systematic and multi-faceted overview facilitates targeted research and modular testing of AV systems. Researchers and developers may benefit from the technical depth and methodological classifications provided, while policymakers and regulators could potentially benefit more from understanding the practical struggles and assessment approaches to make informed decisions about testing requirements. Secondly, it bridges the gap between data-driven and knowledge-driven approaches by demonstrating their complementary roles in identifying edge cases. Last but not least, our analysis of assessment techniques highlights the importance of evaluating the reliability and relevance of identified edge cases. This is a critical aspect, which has often been overlooked in the existing literature.}

In the next sub-section, we first outline the literature search strategy. Then, in Section I-B, we review the related surveys. Section II introduces the relevant notions about edge cases in the \ac{ad} domain; \cref{sec:perception_related,{sec:trajectory_related},sec:knowledge_driven} review the different detection methods for identifying edge cases; \cref{sec:assessment} overviews the techniques and methods for assessing the detection methods and identified edge cases. Finally, in \cref{sec:discussion}, the challenges and future research directions are discussed.

\subsection{Literature Search and Selection Strategy}
Due to the varied and inconsistent terminology used in this domain, we employed a two-level literature identification strategy. At the first level, the group of co-authors collected an initial list of publications based on their expertise in automated driving and edge case detection across both academic and industry contexts. This expert-driven approach was necessary given the fragmented nature of edge case research, which spans multiple disciplines. This initial collection was then expanded through structured database searches using a search string as follows:

\begin{mdframed}[leftline=true, rightline=false,
  topline=false, bottomline=false,
  linewidth=1.5pt, linecolor=gray!60,
  innerleftmargin=6pt, innertopmargin=2pt, innerbottommargin=2pt]
\footnotesize\ttfamily
(``automated vehicle*'' OR ``autonomous vehicle*'' OR ``automated driv*''
 OR ``self-driving car*'' OR ``driverless vehicle*'')\\[2pt]
\textbf{AND} (``edge case*'' OR ``corner case*'' OR ``anomal*''
 OR ``rare event*'' OR ``rare scenario*'' OR ``safety-critical'')\\[2pt]
\textbf{AND} (``detect*'' OR ``identif*'' OR ``recogni*''
 OR ``generat*'' OR ``assess*'' OR ``evaluat*'')
\end{mdframed}

The search was conducted across IEEE Xplore, Scopus, Web of Science, and Google Scholar. Search terms were applied to titles, abstracts, and keywords. Our selection process involved multiple criteria: direct relevance to edge case detection or assessment in automated driving, peer-reviewed articles (or highly cited for pre-prints), and coverage of either perception or trajectory-related edge cases or knowledge-driven methods. The selected publications were validated through cross-validation among the co-authors responsible for the sections of their expertise to ensure completeness, consistency, and balanced representation of methodological approaches and application domains.

\begin{figure}[t]
    \centering
    \includegraphics[width=0.9\linewidth]{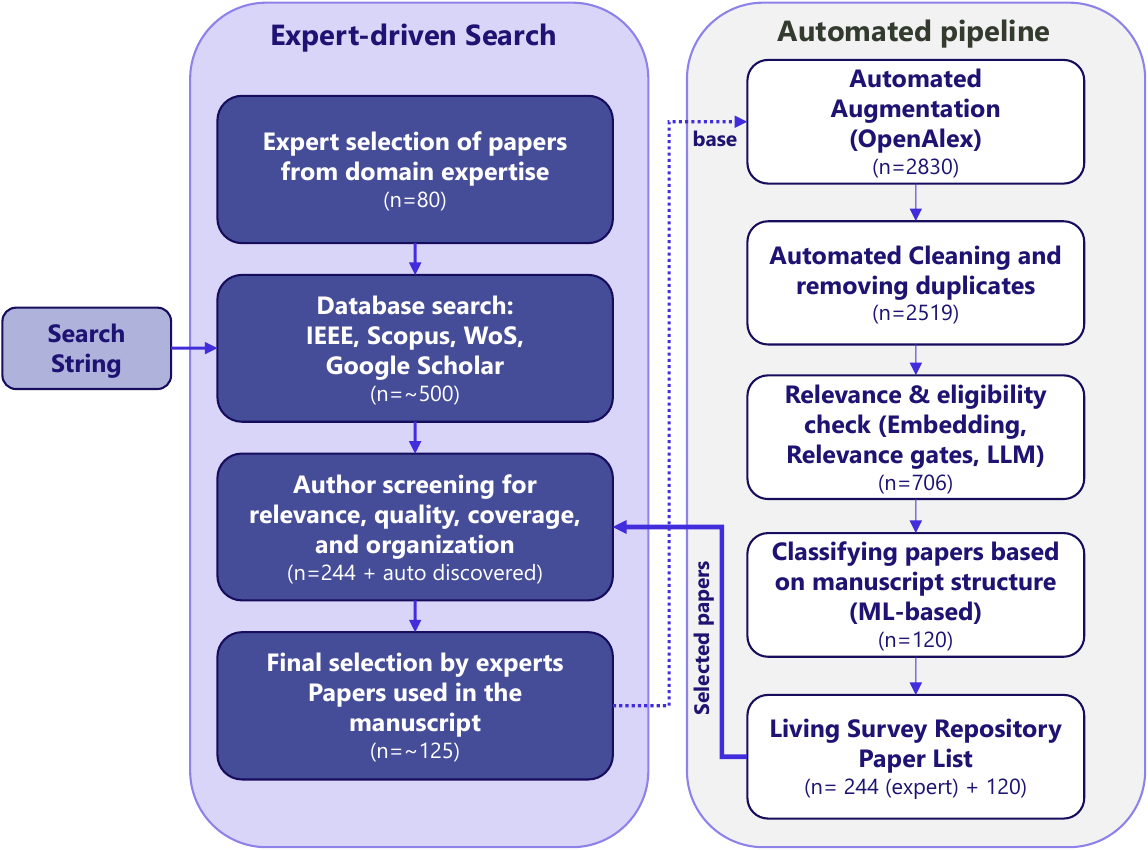}
    \caption{Literature search methodology, including an expert-driven and an automated pipeline.}
    \label{fig:paper_methodology}
\end{figure}

{\color{black} This strategy is accompanied by a computational augmentation to strengthen transparency, methodological rigour, and reproducibility without replacing the expert synthesis. We developed a semi-automated literature discovery and screening pipeline and released it as an open living-survey repository}\footnote{\url{https://github.com/SaeedRahmani/av-edge-case-survey}}{\color{black}. This open and living survey carries the fixed manuscript corpus as a seed, retrieves new records from OpenAlex, removes duplicates and source exclusions, applies embedding-based relevance gates, assigns the survey taxonomy, and retains final human review before appending new papers to the living bibliography. The overall search process and its automated augmentation are summarised in \cref{fig:paper_methodology}.}

\begin{figure}[t]
    \centering
    \includegraphics[width=0.9\linewidth]{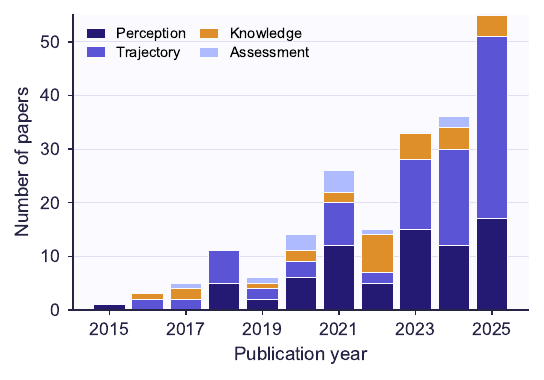}
    \caption{Publication-year trend of edge case detection studies.}
    \label{fig:scientometric}
\end{figure}

{\color{black} A publication-year trend of the identified literature is shown in \cref{fig:scientometric}. The trend indicates that edge-case detection and assessment in automated driving has gained increasing attention in recent years. The growth is especially visible for trajectory-related studies, suggesting that scenario generation, testing, and trajectory-level safety assessment are becoming increasingly active parts of the edge-case literature.}

\subsection{Related Surveys}
\label{sec:related_works}
Current surveys studying edge cases in automated driving can be categorised into two main groups: those focusing on perception-related edge cases and those aiming to address safety-critical scenarios, which can be considered a subset of trajectory-related edge cases. {\cstart \cref{tab:comparison_surveys} presents an overview of previous relevant surveys on edge case or critical scenario detection in the \ac{ad} domain, including a comparison with the current work. In the following, we provide a detailed comparative analysis of these studies.}

\renewcommand{\arraystretch}{1.4}
\begin{table*}[b]
    \caption{A comparison of the current study with the existing surveys related to corner cases and edge cases for automated driving.}
    \centering
    \begin{tabular}{|c|ccc|p{7cm}|}
    \hline
       Study  &  \multicolumn{3}{c|}{Types of edge cases covered} & Analysis limitations \\
       \cline{2-4}
         &  Perception-related & Trajectory-related & Knowledge-driven & \\
\hline
         \textcite{Breitenstein2020SystematizationDriving}
         & \fullcircle & \emptycircle & \emptycircle & Limited to the Classification of \textit{Perception} edge cases; detection methods and assessment techniques are not included \\
         \textcite{Heidecker2021AnDrivingb}
         & \fullcircle & \emptycircle & \emptycircle & Limited to \textit{perception anomalies}; limited coverage of detection methodologies and no discussion on assessment \\
         \textcite{Wang2021AReview}
         & \emptycircle & \halfcircle & \emptycircle & Limited to \textit{safety surrogate measures}; no discussions on the evaluation frameworks \\
         \textcite{Zhang2022Finding}
         & \emptycircle & \halfcircle & \emptycircle & Limited to \textit{safety-critical} scenarios, which are a subset of the wide spectrum of edge cases;  \\
         \textcite{Bogdoll2022AnomalySurvey}
         & \halfcircle & \halfcircle & \emptycircle & Focusing on \textit{anomalies}; limited coverage of trajectory-related cases; no discussions on knowledge-driven approaches;  not covering assessment techniques\\
\hline
         This work
         & \fullcircle & \fullcircle & \fullcircle & Comprehensive coverage across all edge case types, detection methods, and assessment techniques, metric, and datasets \\
\hline
         \multicolumn{5}{|l|}{\fullcircle~: Aspect covered \quad \halfcircle~: Aspect partially covered \quad \emptycircle~: Aspect not covered} \\
    \hline
    \end{tabular}
    \label{tab:comparison_surveys}
\end{table*}

Within the first group of surveys, \textcite{Breitenstein2020SystematizationDriving} provide a systematic categorisation of visual \textit{perception} corner cases based on detection complexity, organising them hierarchically from pixel-level issues to scenario-level temporal patterns. Building on this foundation, \textcite{Heidecker2021AnDrivingb} extend corner case systematisation to multi-sensor perception environments and introduce a layer-based framework (sensor, content, temporal, and method layers) to address sensor-specific manifestations and data fusion challenges. While both studies offer valuable perception taxonomies, they focus on categorising corner cases rather than reviewing detection methodologies, remain limited to perception-related edge cases, and do not explore knowledge-driven methods and emphasise data-driven approaches.

Focusing on trajectory-related edge cases, \textcite{Wang2021AReview} provided an extensive review of safety surrogate measures in evaluating AV safety, covering metrics such as time-to-collision, brake threat number, and other conflict-based indicators. While comprehensive in its treatment of surrogate safety metrics, their survey is methodologically narrow and focuses exclusively on predefined metric-based approaches without addressing machine learning-based detection methods, probability estimation techniques, or system-challenging approaches. Moreover, their scope is limited to safety-critical scenarios.  \textcite{Zhang2022Finding} discussed methods for identifying safety-critical scenarios and provided a structured overview of verification and validation approaches for AVs. Their survey covers scenario generation techniques and testing methodologies but also remains limited to safety-relevant cases.

\textcite{Bogdoll2022AnomalySurvey} represent the most comprehensive existing survey by reviewing anomaly detection methods across camera, lidar, radar, multimodal, and object-level classifications. Their methodological coverage includes various detection approaches applicable to these anomaly classes and provides technical depth in anomaly detection algorithms. However, several critical limitations constrain their contribution: (1) primary focus remains on perception-related anomalies with only brief treatment of trajectory-related cases; (2) coverage is limited to anomalies; (3) assessment techniques for evaluating detection methods are not systematically addressed; and (4) knowledge-driven approaches are absent.

Overall, while these surveys make valuable contributions to specific aspects of edge case research, they exhibit limitations in scope, methodological coverage, or practical applicability. They primarily focus on specific types of edge cases, such as perception-related \textit{anomalies} or \textit{safety-critical scenarios}. They also lack a consistent mapping of detection methods to their assessment techniques and metrics. Our survey distinguishes itself by comprehensively reviewing and categorising various classes of edge cases based on their affected subsystems and mapping them to relevant detection methods. We also introduce a new category of edge case identification methods---knowledge-driven edge cases, largely overlooked in the literature. We go beyond reviewing the edge case detection methods and investigate the techniques and metrics for assessing the relevance of identified edge cases and their detection methods. With these contributions, we aim to offer a more comprehensive and structured approach to understanding, detecting, and evaluating edge cases for AV.
{\cstart \section{Preliminaries and Research Scope}}
\label{sec:preliminaries}

\subsection{Definition of an Edge Case}
\label{sec:definition}
The term ``edge case'' carries varied definitions across fields  \cite{Vater2023ADriving}. In \ac{ad}, ``edge cases'' are mainly used for safety testing and validation, but it is hard to find a unified definition for the term. This absence of a universal definition and quantitative thresholds reflects several fundamental challenges. Statistical rarity thresholds are inherently context-dependent, varying significantly based on the specific AV system capabilities, operational design domain, regulatory requirements, and organisational safety objectives. While ISO 21448 (SOTIF) addresses unknown hazardous scenarios, it does not provide explicit quantitative definitions for edge cases, instead emphasising systematic identification and mitigation of hazardous behaviour.

\textcite{Koopman2019Credible} define an edge case as ``a rare situation that will occur only occasionally but still needs specific design attention to be dealt with in a reasonable and safe way. The quantification of `rare' is relative and generally refers to situations or conditions that will occur often enough in a full-scale deployed fleet to be a problem but have not been captured in the design or requirements process.''

\textcite{Vater2023ADriving} expanded the \textcite{Koopman2019Credible}'s definition by introducing two concepts, ``novel'' and ``boundary case.'' ``novelty'' refers to the new additions to driving environments (which are not necessarily ``rare''). For instance, the introduction of a new mobility means (such as e-scooters) could be considered an edge case at the time of introduction. The concept ``boundary case'' intends to distinguish an \textit{edge case} from a \textit{corner case}. {\cstart They propose that an edge case is a boundary case of \textit{one parameter}, while corner cases refer to situations where the \textit{combination} of normal operational parameters may lead to a rare situation. This distinction clarifies that edge cases involve single parameters reaching extreme values, such as exceptionally low visibility conditions, whereas corner cases emerge from the \textit{simultaneous} occurrence of multiple individually normal conditions: moderate rain combined with curved roads, dense traffic, and sudden braking events that together create a rare and challenging scenario.}

{\cstart In this study, we use the definition provided by \textcite{Vater2023ADriving} and define the term edge case as: 

\textit{An edge case is a novel or rare situation that still needs specific design attention to be dealt with in a reasonable and safe way. The quantification of rare is relative and generally refers to situations or conditions that will occur often enough in a full-scale deployed fleet to be a problem.}}

{\cstart Despite the distinction between edge cases and corner cases, our survey encompasses methods and studies on both edge and corner case detection due to their frequent interchangeable use in the literature and their shared importance in identifying challenging scenarios for \acp{av}.
It is worth noting that in some studies, the term ``anomaly'' is also used as an alternative word for ``edge case,'' but this terminology is not accurate. While anomalies involve data deviating significantly from expected parameters, edge cases encompass broader scenarios, which also include normal data combined in rare configurations (for instance, high speed + low visibility + sharp turns). Therefore, anomalies are often a subset of edge cases. However, not all anomalies are edge cases either. For example, an unusual object being far away from the drivable area might be considered an anomaly but not an edge case since it will not lead to a challenging situation for the \ac{av}.}

\subsection{Taxonomies of Edge Cases and Identification Methods}\label{sec:identification_classes}
This section examines the main approaches to categorizing edge cases to provide a foundation for better understanding of our approach in this survey.

\subsubsection{Detection vs.\ Generation}
Edge cases can be achieved or identified through two main mechanisms: detection and generation. The former method identifies the edge cases during the analysis of existing datasets and usually represents unusual or infrequent patterns in the data. In contrast, generated edge cases are created through simulations or synthetic data generation techniques to anticipate rare conditions not present in the training data. In this survey, we use the term edge case ``identification'' to cover both methods.

\subsubsection{Data-driven vs. Knowledge-driven}
Edge-case identification methods can be classified as data-driven or knowledge-driven techniques. Data-driven techniques identify the edge cases by analysis of large datasets by leveraging statistical methods or machine learning algorithms. Knowledge-driven edge cases are managed by integrating expert knowledge and rely on the understanding and anticipation of potential problems based on domain expertise, historical precedents, and theoretical models. It emphasizes a proactive stance where systems are equipped with the necessary logic to handle known edge cases even before they are encountered. These two approaches offer a supplementary strategy for tackling edge cases.

\subsubsection{Online vs.\ Offline Detection}
Edge case detection can be applied online or offline. Online methods continuously analyse data in real-time and enable swift detection and response before issues escalate. They also help with avoiding large data collection by focusing on unusual or interesting scenarios. In contrast, offline methods involve the detection of edge cases from historical data, simulations, or controlled experiments. This approach allows for a more thorough and detailed examination of the data, enabling the identification of subtler patterns or anomalies that might not be apparent in real-time analysis. Offline methods provide the opportunity to refine and optimise algorithms more thoroughly based on comprehensive insights gained from the data.
 
\subsubsection{Root Causes and System Effects}
Edge cases can be studied based on their roots in or effects on different modules of \iac{av}, including perception, decision-making and planning, and control. This perspective is important as addressing each type of edge case requires investigation or improvement of its relevant subsystem. Perception-related edge cases may involve sensory distortions or misclassification of objects, while decision-making and planning-related edge cases arise in the strategic and tactical layer of the system, where unforeseen scenarios disrupt the formulation of action plans. These might include dynamic obstacles or sudden changes in surrounding environments. Control-related edge cases involve challenges in executing the planned actions effectively, possibly due to mechanical limitations or external interferences.

\subsection{A Hierarchical Approach to Methods Classification}\label{subsect:categorization}
{\cstart The multi-faceted variety of perspectives on edge cases underscores the complexity of identifying and addressing them. In addition to the classification of identification methods, edge cases themselves can be grouped into different types, including safety-critical scenarios (e.g., sudden pedestrian intrusions), statistical anomalies (e.g., unusual object behaviours), performance degradation cases (e.g., reduced perception accuracy), and environmental extremes (e.g., severe weather events). However, these groups often overlap; a single scenario may simultaneously be safety-critical, statistically rare, and environmentally extreme. This overlapping nature, combined with the inherently unpredictable and novel characteristics of edge cases, makes creating detailed classification systems difficult and potentially unhelpful. Accordingly, this survey adopts a methodologically focused approach that focuses on reviewing and categorising detection methodologies and assessment techniques rather than edge cases themselves. Our classification system is structured on two primary levels:}

The first level takes a practical perspective by categorising the detection methods according to different modules of an \ac{av}. We distinguish between \textit{perception-related} methods and those concerned with planning, decision-making, and control, which we collectively refer to as \textit{trajectory-related} edge cases. These two classes of methods fall under the umbrella of data-driven methods. Next, we introduce ``knowledge-driven'' methods, which leverage expert insights, potentially complemented by data analysis when available. Knowledge-driven approaches can complement and encompass both trajectory-related and perception-related cases and potentially identify edge cases that have not yet been observed in the datasets. In the second level, we further classify identification methods based on their underlying methodological approaches, such as generative techniques, confidence scores, and machine learning approaches. 

\begin{figure*}[t]
    \centering
    \includegraphics[width=0.75\linewidth]{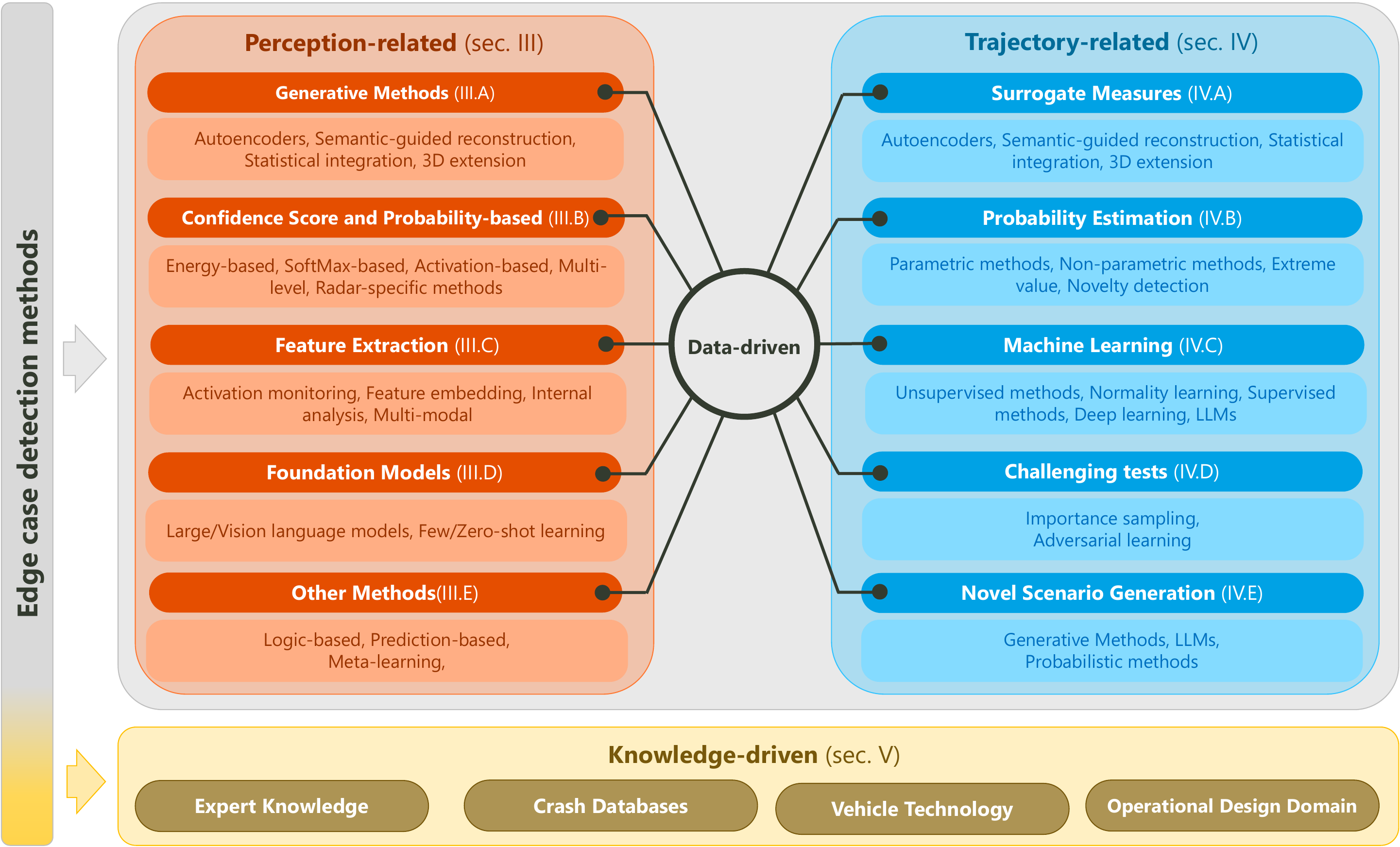}
    \caption{The taxonomy of edge case detection methods proposed in this study (The numbers in front of methods indicate the section number discussing them.). }
    \label{fig:edge_case_types}
\end{figure*}

This dual-level classification enables modular testing of subsystems, supports targeted resource allocation, and helps with identifying suitable detection methods considering their strengths and weaknesses. It also provides a foundation for comparing detection methods, identifying research gaps, and guiding future work. \cref{fig:edge_case_types} provides a taxonomy of different edge case detection methods. 

{\cstart
\subsection{Sensory Technologies and Processing Paradigms}
Edge case detection depends heavily on the sensing technologies used, making it essential to understand sensor limitations. While a detailed sensor analysis is beyond this survey's scope, typical AV configurations combine diverse modalities to gather complementary data. Cameras provide rich semantic details but struggle with poor lighting, glare, and occlusions. LiDAR offers precise 3D geometry regardless of lighting but degrades with reflective surfaces or long-range sparsity. Radar reliably measures range and velocity in bad weather, though it can suffer from lower angular resolution and ghost returns. Ultrasonic sensors handle low-speed, close-range perception, while GNSS and IMU track position and motion despite potential drift. Together, these sensors offset each other's limitations to create a comprehensive detection strategy \cite{wang2019multi}.

Since many edge cases are modality-conditioned, fusion and processing paradigms are important when addressing edge cases. Rule-based methods are interpretable but rigid, while learning-based methods capture complex patterns but struggle with explainability and out-of-distribution scenarios. Hybrid approaches that blend physical models with data-driven learning offer greater robustness, especially when combined with temporal analysis across multiple frames. Understanding the trade-offs across sensing and processing approaches guides the selection of appropriate detection methods. Comprehensive discussions are covered in related reviews \cite{wang2019multi, marti2019review}.}

\subsection{Survey Scope and Related Domains}
This survey focuses on edge case detection methods for operationally and environmentally driven scenarios in automated driving. Yet, it is important to acknowledge the broader landscape of anomaly detection research, particularly in the cybersecurity domain. There exist overlaps between edge case detection and cybersecurity threat identification methods, as both domains employ similar analytical approaches, including machine learning-based anomaly detection, statistical outlier identification, and behavioural pattern recognition \cite{kim2021cybersecurity, yousseef2024autonomous}.

Despite these similarities, important distinctions must be acknowledged. Not all cybersecurity incidents can be considered edge cases because many represent intentionally engineered adversarial manipulations rather than rare or safety-critical events. That said, cyber-attacks can induce operational anomalies that manifest as edge cases. For instance, spoofed sensor signals may result in perception errors, or manipulated communication streams can lead to abnormal trajectories. In this survey, we remain focused on detection methodologies for identifying these \textit{operational manifestations} of edge cases rather than on detecting or attributing the underlying attacks themselves. 
Accordingly, we position security-induced edge cases as a complementary but distinct research domain and refer the readers to existing reviews of cybersecurity methods for connected and automated vehicles \cite{kim2021cybersecurity, girdhar2023cybersecurity}.
\section{Perception-related edge cases}\label{sec:perception_related}
Perception forms the foundation of \ac{av} functionality by providing environmental awareness for planning and control. Perception-related edge cases arise where unusual environmental conditions, sensor limitations, or unexpected object appearances challenge the AV's ability to accurately detect, classify, or interpret its surroundings. The methods used for detecting perception-related edge cases are frequently considered to belong to the broader category of out-of-distribution (OoD) detection techniques. While existing surveys exist for generic OoD detection \cite{Yang2021GeneralizedSurvey, Salehi2022AChallenges, ChengChih-HongandLuttenberger2023RuntimePerception}, these are insufficient for automated driving's multi-object scenarios. This survey focuses specifically on methods relevant to automated driving.

We divide these methods into five categories, namely reconstructive and generative methods, confidence scores, feature extraction, foundation models, and other techniques that go beyond these classes \cite{Breitenstein2020SystematizationDriving, Bogdoll2022AnomalySurvey}. \cref{tab:summary_perception} provides an overview of the different categories of methods for detecting perception-related edge cases in \ac{av} systems. In the next subsections, we take a deeper look into these studies and synthesize them.

\begin{table*}[b]
    \centering
    \caption{Overview of the different categories for detecting perception-related edge cases.}
    \begin{tabular}{lll}
    \hline
    Method & Publications & Characteristics \\
    \hline
    Reconstructive and Generative & \cite{Vojir2021RoadCoupling}, \cite{Cai2020Real-timeSystems}, \cite{ramakrishna2022efficient}, \cite{Cultrera2023LeveragingDetection},  & Train a machine learning model on the input data \\ (\cref{sec-perc:generative}) & \cite{DiBiase2021Pixel-wiseScenes}, \cite{Stocco2020MisbehaviourSystems},\cite{Masuda2021TowardAutoencoder}, \cite{Lis2019DetectingResynthesis} & Distinguish anomalies based on reconstruction errors of the trained model \\
    \hline
    Confidence Scores & \cite{Liu2020Energy-basedDetection}, \cite{ Wang2021CanKnow}, \cite{ Liang2018EnhancingNetworks}, \cite{Hsu2020GeneralizedData}, \cite{Sun2022DICE:Detection}, \cite{katz2022training}, \cite{Bai2023FeedDetection}, \cite{Du2022Unknown-AwareWild},  & Detects edge cases based on scores \\ (\cref{sec-perc:scores}) & \cite{Lin2021Mood:Detection}, \cite{noguchi2024road}, \cite{Wang2021RadarTransformers}, \cite{Ryu2018DetectingTracking}, \cite{Kopp2023TacklingPointnet++}, \cite{Kraus2020UsingRadar}, \cite{Sun2021ReAct:Activations} & Scores are often computed based on the last layer\\
    \hline
    Feature Extraction & \cite{Cheng2018RuntimePatterns}, \cite{Henzinger2020OutsideNetworks}, \cite{Lukina2021IntoNetworks}, \cite{Cheng2021Provably-robustPatterns}, \cite{Hashemi2021Gaussian-BasedNetworks}, \cite{Hashemi2023RuntimeNetworks},   & Extract intermediate data from the NN; Features are often post-processed \\ (\cref{sec-perc:features}) &  \cite{Sastry2020DetectingMatrices}, \cite{Morteza2022ProvableDetection}, \cite{Wilson2023SAFE:Detection}, \cite{Sun2022Out}, \cite{Luan2021Out-of-DistributionFactor}, \cite{Sun2020Real-timeImages}& Detect edge cases based on these processed features \\
    \hline
    Foundation Models & \cite{ elhafsi2023semantic}, \cite{loukas2025evaluation}, \cite{zou2025few}, \cite{li2024few},  & Leverage large pre-trained language/vision-language models \\ (\cref{sec-perc:foundation}) & \cite{chen2025insight}, \cite{shriram2025towards} & Few- and zero-shot adaptation with minimal training data  \\
    \hline
    Other methods & \cite{Balakrishnan2021PerceMon:Systems}, \cite{Bolte2019TowardsDriving}, \cite{Liu2018FutureBaseline}, \cite{Lv2021LearningNetwork} & Monitoring of logic properties, future frame prediction, and meta-learning \\ (\cref{sec-perc:others}) &&\\
    \hline
    \end{tabular}
    \label{tab:summary_perception}
\end{table*}

\subsection{Reconstructive and Generative Methods}\label{sec-perc:generative}
Reconstructive and generative methods emerged as promising strategies for detecting edge cases in \ac{av} perception systems. Reconstructive methods analyse and rebuild the sensor data or images to identify anomalies or unexpected scenarios that do not match the training data. The idea of reconstructive methods is that recreating an image containing an anomaly leads to a higher reconstruction error. These methods often use techniques such as autoencoders to reconstruct inputs and measure deviations from the original data \cite{an2015variational, torabi2023practical}. Alternatively, generative methods create new data instances that mimic realistic but unseen driving scenarios. These methods often involve generating new data and using a discriminator to evaluate deviations from expected distributions \cite{wang2018generative}. The similarity between reconstructive and generative methods lies in their use of \textit{deviation metrics} to detect anomalies \cite{Breitenstein2020SystematizationDriving}. Accordingly, we review them as one category in this subsection.

A common strategy within this domain is to combine semantic segmentation with reconstruction techniques. \textcite{Vojir2021RoadCoupling} and \textcite{Lis2019DetectingResynthesis} proposed methods that first apply semantic segmentation to an input image and then use the resulting features or segmentation map to reconstruct the original image. Areas with significant differences between the original and reconstructed images are flagged as potential anomalies. \textcite{DiBiase2021Pixel-wiseScenes} extended this concept by adding a synthesis network between the segmentation and anomaly prediction steps to improve anomaly detection without compromising segmentation accuracy.

Autoencoders are another popular technique in reconstructive approaches. \textcite{Cai2020Real-timeSystems} explored the use of \ac{VAE} and \ac{SVDD} for anomaly detection. The \ac{VAE} method involves training on known safe data, sampling points from the latent space during runtime, and using the reconstruction error to detect anomalies. The \ac{SVDD}, trained similarly, calculates the distance of predictions to the centre of a hypersphere, with larger distances indicating anomalies. \textcite{ramakrishna2022efficient} advanced this idea by employing a $\beta$-\ac{VAE} with a disentangled latent space, allowing for more interpretable results and reasoning about the model's decisions. Other researchers, such as \textcite{Cultrera2023LeveragingDetection} and \textcite{Stocco2020MisbehaviourSystems}, investigated various autoencoder architectures, including \ac{CAE} and deep fully connected autoencoders, often combining them with statistical techniques like fitting reconstruction errors to a Gamma distribution for more reliable anomaly detection. While most of these methods focus on 2D image data, some researchers have expanded the approach to other data types. \textcite{Masuda2021TowardAutoencoder} applied a similar autoencoder-based reconstruction technique to 3D point cloud data for anomaly detection.

While both reconstructive and generative methods have demonstrated effectiveness in controlled environments, their real-world application faces several challenges. The primary concern is their proneness to false positives and false negatives that could lead to unnecessary interventions or missed critical scenarios. These methods rely on the unproven assumption that anomalies cause higher reconstruction errors. This assumption may be problematic in safety-critical applications like automated driving. Moreover, their effectiveness depends heavily on the quality and coverage of training data. The computational demands of these methods are another concern, especially for real-time applications. In addition, there is an ongoing debate about the interpretability of these models, which is essential for safety-critical applications. Future research in this area should focus on improving the interpretability, efficiency, and reliability of these methods. Additionally, there is a growing interest in developing these methods to work more effectively with multi-modal sensor data, reflecting the complex sensor suites used in modern \acp{av}. 

\subsection{Probability-based Methods and Confidence Scores}\label{sec-perc:scores}
These methods quantify the uncertainty of perception outputs. Probability-based approaches estimate the likelihood of an observation under a learned distribution, while confidence scores quantify the level of certainty with which a perception system identifies and classifies objects and scenarios \cite{Yang2021GeneralizedSurvey,Salehi2022AChallenges}.

Among different methods, the energy score \cite{Liu2020Energy-basedDetection} provides a basic yet popular approach. Energy-based methods involve mapping each input to a scalar energy value, where lower energy corresponds to a higher likelihood of being \ac{id}. \textcite{Wang2021CanKnow} extended energy scores to multi-label classification networks with the JointEnergy method, which aggregates label-wise energy scores to capture joint uncertainty across all labels. \textcite{katz2022training} focused on leveraging unlabelled or ``wild'' data to improve OoD detection. They presented WOODS (Wild OoD detection sans-Supervision), which explicitly optimises for a level-set estimation based on the energy function. The authors formulate a constrained optimisation problem to ensure the model can correctly classify normal data while identifying anomalies. 
Building upon WOODS, \textcite{Bai2023FeedDetection} proposed \ac{SCONE}, which improves on WOODS by considering both semantic shifts and covariate shifts.

Alternative techniques focus on the softmax probabilities in the last layer of the \acp{nn} and combines it with temperature scaling and small perturbations of the input. \textcite{Liang2018EnhancingNetworks} proposed the Out-of-DIstribution detector for Neural Networks (ODIN), which improves the separability of score distributions between \ac{id} and OoD images. \textcite{Hsu2020GeneralizedData} further developed ODIN by addressing the limitation of needing specific \ac{ood} data for parameter tuning by modifying input preprocessing and decomposing confidence scoring.

Others have explored activations and weights within the neural networks. \textcite{Sun2022DICE:Detection} introduced Directed Sparsification (DICE), a method that identifies and utilises only the most salient weights based on their contribution to the model's output. By ranking weights and selectively using the most important ones, DICE reduces output variance for \ac{ood} data, resulting in a sharper and more separable output distribution.
Complementing DICE, \textcite{Sun2021ReAct:Activations} proposed Rectified Activations (ReAct) truncated high activation patterns triggered by \ac{ood} data. By setting an upper limit on activations, ReAct reduces the model's overconfidence on \ac{ood} inputs without modifying the network's training process. This post-hoc technique is supposed to generalise across different network architectures and \ac{ood} detection scores.

Extending \ac{ood} detection to video data, \textcite{Du2022Unknown-AwareWild} introduced the Spatial-Temporal Unknown Distillation (STUD) framework. STUD leverages spatial and temporal information from video data to distill unknown objects, which are then used to regularize the model's decision boundary. STUD employs an energy-based uncertainty regularization loss to shape the uncertainty space between \ac{id} objects and distilled unknown objects. 
Addressing the computational efficiency of \ac{ood} detection, \textcite{Lin2021Mood:Detection} proposed Multi-Level Out-of-Distribution Detection (MOOD). MOOD incorporates multiple \ac{ood} detectors at various network depths. This method allows for early detection of less complex \ac{ood} examples and deeper processing of more complex cases, balancing computational costs with detection performance. \textcolor{black}{More recent AV-specific work extends confidence-based reasoning to open-set road-obstacle detection: \textcite{noguchi2024road} use unknown-objectness scores to identify road obstacles outside closed-set detector labels.}

Previously reviewed studies have focused on camera-based images. In radar-collected data, ghost objects can appear when a wave detecting an object is reflected by another surface, leading to erroneous detection. To address this issue, \textcite{Ryu2018DetectingTracking} introduced a multi-layer perceptron trained on key characteristics of ghost targets, such as track lifetime and velocity differences.
\textcite{Wang2021RadarTransformers} proposed a multimodal transformer network to detect radar ghost targets by fusing radar and lidar data. They employ self-attention mechanisms to facilitate information exchange between radar points and cross-modal attention to infuse information from surrounding lidar points. \textcite{Kraus2020UsingRadar} introduced PointNet++ to classify radar detections into real objects, ghost objects, and background clutter and showed effective reduction of false positives caused by ghost images. \textcite{Kopp2023TacklingPointnet++} improved PointNet++ for clutter detection by introducing new network variants and an innovative automatic label generation method. They also developed an accumulation-aware downsampling technique to enhance processing efficiency. These diverse methods demonstrate ongoing efforts to enhance the robustness and accuracy of radar-based perception systems in \acp{av}.

In summary, while energy-based scores effectively differentiate \ac{id} and \ac{ood} data, advanced methods that incorporate input perturbation and temperature scaling further refine this distinction. Additionally, frameworks utilising unlabelled data and addressing both semantic and covariate shifts have significantly advanced the field, offering robust solutions for comprehensive anomaly detection. Despite their potential, these methods often only consider limited information as they mostly focus on the output of a network.
An approach trying to mitigate this risk is MOOD \cite{Lin2021Mood:Detection}, which uses several different confidence scores at different depths of the model.
Further research can focus on enriching confidence scores with more information or combinations with other methods. Also, increasing the efficiency and generalisability of confidence score methods, reducing their reliance on extensive labelled datasets, and their integration into multimodal data sources are other open research areas. Improving the interpretability of these scores will be essential for developing more reliable and safer \ac{av} perception systems. 

\subsection{Feature and Activation Value Extraction} \label{sec-perc:features}
These methods use \acp{nn} to extract and analyse features from input data to identify potential edge cases. These features can include a perception system's intermediate computations or activation values. By examining these internal representations, researchers aim to understand how \acp{nn} interpret data and make decisions, potentially revealing patterns that indicate edge cases or out-of-distribution scenarios.

\textcite{Cheng2018RuntimePatterns} suggested monitoring the internal behaviour of a \ac{nn} as it contains more information about the computations and potentially helps obtain a deeper understanding of the output. They used the activation values of neurones of a hidden layer to construct patterns from these values with a binary decision diagram. If a previously unseen pattern is observed at runtime, the example is assumed to be an anomaly. \textcite{Henzinger2020OutsideNetworks}
proposed collecting the activation patterns and clustering them into separate boxes. Normal data is assumed to always fall within one of these boxes. If the activation values of the monitored layer are not within the box, the input is said to be an edge case. \textcite{Cheng2021Provably-robustPatterns} improved monitoring algorithms in \textcite{Henzinger2020OutsideNetworks} by using verified safe ranges instead of using only feature vectors. These safe ranges are calculated to include all possible small changes in the input images. This approach guarantees that any changes within these bounds won't affect the network's safety, reducing false alarms and increasing trust in the system.

\textcite{Lukina2021IntoNetworks}
proposed a framework that operates in parallel with a neural network, using an adaptive quantitative monitor that interacts with a human authority to obtain correct labels for novel inputs. The framework includes mechanisms for automatic switching between monitoring and adaptation modes based on runtime statistics, allowing for the incremental adaptation of both the monitor and the neural network. 

\textcite{Hashemi2021Gaussian-BasedNetworks}
suggested fitting a Gaussian distribution to the activations of each neurone for each class. Then, an interval containing a certain percentile of the safe data of the neurone activation values is calculated. During runtime, sufficient neurones should fall within their respective interval for the monitor to consider the image in-distribution. They extended the Gaussian monitor to object detection by computing a Gaussian for each neurone independent of the predicted classes \cite{Hashemi2023RuntimeNetworks}. Similarly, \textcite{Sastry2020DetectingMatrices} focused on the activation values of an intermediate layer.
They computed a Gram matrix for these layers, which is the matrix of activation values multiplied by its transposed form.
This way, pairwise relations between the channels can also be observed.  
They also considered higher-order gram matrices for more detailed activation value connections. At runtime, they compare these correlations to those observed in training data. Samples showing significantly different behaviour are classified as edge cases. 

\textcite{Morteza2022ProvableDetection}
proposed a new scoring function for detecting \ac{ood} data called Gaussian mixture-based energy measurement (GEM). GEM models the feature space as a class-conditional multivariate Gaussian distribution, using a Gaussian generative model to derive a new \ac{ood} scoring function. They show it outperforms previous methods such as the energy score and maximum Mahalanobis distance by a significant margin.

\textcite{Sun2022Out} leveraged internal feature embeddings, such as layer activation values, to compute distances to k-nearest neighbours. They proposed using the penultimate layer for this purpose. Inputs with distances exceeding a certain threshold were classified as \ac{ood}. This approach is advantageous as it requires no assumptions about neurone value distributions and remains model-agnostic and \ac{ood}-agnostic.

In addition to using neural networks to analyse the activation values and patterns, it is possible to utilise them to detect edge cases on feature vectors.
\textcite{Wilson2023SAFE:Detection}
suggest focusing on residual convolution layers with batch normalisation. They extract vectors from these layers and train a multi-layer perceptron to distinguish (adversarially) perturbed images. \textcite{Sun2020Real-timeImages} proposed RFNet. The method utilises an RGB sensor and additional depth estimation to predict obstacles on the road.
Two ResNet-18 backbones compute features from an RGB image and a depth image. The computed feature maps are fused, and the resolution of the fused maps is upsampled to the original size to use for the prediction of potentially dangerous objects on the road.

To conclude, these methods can provide a deeper understanding of the neural networks by examining intermediate computations, which increases the interpretability of the outputs. However, the application of these techniques primarily focuses on classification tasks, limiting their use in more complex scenarios such as object detection. While some methods have been adapted for object detection, there remains a need for more comprehensive solutions that can handle the diverse and dynamic nature of real-world driving environments. Future research may focus on further improving the interpretability of the extracted features and activation patterns. 

{\cstart 
\subsection{Foundation Model and Adaptive Learning}\label{sec-perc:foundation}
Foundation models and adaptive learning promise a significant methodological advancement in edge case detection. Foundation models are large neural networks pre-trained on diverse datasets and have proven to adapt to multiple downstream tasks. Large language models (LLMs) and vision-language models (VLMs) are examples of such models. The broad pre-training of foundation models is argued to enable reasoning capability about novel scenarios with minimal task-specific training \cite{elhafsi2023semantic}. This has introduced a paradigm shift in edge case detection by identifying semantic anomalies, which are \textit{system-level} failures caused by unusual combinations of individually normal elements rather than \textit{component-level} malfunctions. Unlike traditional perception-based anomaly detection that focuses on OoD visual patterns, LLM-based approaches leverage contextual reasoning capabilities to identify scenarios where typical objects appear in misleading or unexpected configurations, such as inactive traffic lights being transported on trucks or stop signs appearing on roadside billboards \cite{loukas2025evaluation, elhafsi2023semantic}. \textcite{elhafsi2023semantic} pioneered this approach by converting robot observations into natural language descriptions processed through carefully engineered prompts.

Combining foundation models with adaptive learning techniques, such as few- and zero-shot learning, has further enhanced their capability to handle anomalous inputs with minimal training data \cite{zou2025few, li2024few}. Manifold-enhanced frameworks decouple high-dimensional LLM textual spaces into low-dimensional representations, which enables commonsense reasoning within domain-specific decision manifolds while maintaining external self-correction databases for continual learning \cite{zou2025few}. Cross-attention similarity mechanisms enable learning from AV testing scenario spaces and transform edge case testing from probabilistic sampling to deterministic optimisation when training budgets are constrained \cite{li2024few}. Vision-language models have demonstrated zero-shot hazard detection capabilities through integrating semantic and visual representations, which enables identification of novel hazardous objects without extensive retraining \cite{chen2025insight, shriram2025towards}. These developments position foundation model-based approaches as a complementary methodology, particularly valuable for proactive identification of edge cases that may not appear in training datasets.
}

\subsection{Other Methods}
\label{sec-perc:others}
Due to the diverse nature of perception-related edge cases and their detection methods, it is not feasible to fit all detection methods into certain groups. This section overviews some of the most popular and promising instances of such methods that are not fitted to the overviewed categories of methods.
\textcite{Balakrishnan2021PerceMon:Systems} proposed
Spatio-Temporal Quality Logic (STQL) to express correctness properties for perception algorithms, which allows them to monitor complex scenarios involving object detection and tracking. The tool integrates with the CARLA simulator and the Robot Operating System (ROS) \cite{quigley2009ros} for real-time monitoring of perception modules.

\textcite{Bolte2019TowardsDriving}
proposed an image prediction approach that anticipates future frames and calculates prediction errors, which are then analysed in conjunction with semantic segmentation results to generate a corner case score. Similarly, \textcite{Liu2018FutureBaseline} proposed an anomaly detection method based on predicting future frames in video sequences. This approach leverages the difference between the predicted future frame and the actual frame to identify anomalies. Their model incorporates both spatial constraints (intensity and gradient losses) and temporal constraints (optical flow loss) to ensure high-quality predictions.

\textcite{Lv2021LearningNetwork} suggest using prior knowledge and learning normal dynamics by capturing common patterns through meta-prototypes. The idea is to combine meta-learning and prototype learning to effectively generalise across different scenarios and datasets. Their experimental results show that this approach improves anomaly detection performance, particularly in complex and dynamic environments.

\subsection{Conclusion}
This section organised the perception-related edge case detection methods  into five categories: reconstructive and generative methods, confidence scores, feature extraction, foundation models, and other techniques, including logic-based, prediction-based, and using prior knowledge. These categories differ in where they interface with the perception pipeline. Generative and reconstructive methods typically operate on the system's input, feature extraction engages with its internal representations, confidence scores focus on its output, and foundation models reason over semantic descriptions of the scene. Some methods also incorporate temporal information, such as future frame prediction. Given the diversity of failure modes, a combination of these approaches is likely necessary to ensure safe deployment.
Further research should extend the application domain of these methods beyond classification tasks.
It is essential to verify if such methods can directly be applied to object detection or which changes are required.
Furthermore, the explainability of monitoring techniques should be investigated as obtaining an understanding of why the system alerted of an edge case is important for improving the \ac{av} and the detection mechanism.
\section{Trajectory-related edge cases}
\label{sec:trajectory_related}

Trajectory-related edge cases challenge an AV's planning, decision-making, and control capabilities and are often directly or indirectly linked to the vehicle's trajectory calculation, prediction, or execution across various driving tasks \cite{Laxhammar2014Online, Morris2008A, Rosch2022Space}. These tasks may include longitudinal driving, lateral manoeuvres, and turning manoeuvres at intersections. Edge cases can manifest as sudden changes in road conditions, unpredictable movements of other road users, or system failures in executing planned trajectories. Trajectories (a finite temporal sequence of data points) \cite{Laxhammar2014Online, Heidecker2019Towards}, often form the basis for understanding and detecting these scenarios.

Five different categories of detection methods for trajectory-related edge cases are identified: using surrogate measures of safety, such as \ac{ttc}; looking into probability estimation of trajectory parameters (e.g., longitudinal or lateral velocities); applying machine learning methods; identifying challenging situations for the system under test (e.g., from previous knowledge or assumptions); and novel scenario generation using machine learning techniques. \Cref{tab:trajectory based} provides an overview of these methods by highlighting the relevant literature and main characteristics. We detail these methods and their relevant studies in the following subsections.

\begin{table*}[t]
    \centering
    \caption{Overview of the different categories for the trajectory-based edge case detection.}
    \label{tab:trajectory based}
    \begin{tabular}{lll}
        \hline
        Method & References & Characteristics \\ \hline
        
        Deriving edge cases using metrics &
        \cite{ Wang2021AReview}, \cite{Hayward1972NearMiss}, \cite{ ke2017cost}, \cite{ Brannstrom2008Situation},    &
        Proven to be successful in predicting accident rate \\
        (\cref{subsec:metrics}) & \cite{ Asljung2016Comparing}, \cite{ Ponn2020Identification}, \cite{Yu2020Method}, \cite{ Wang2018Traffic}, & Many metrics are possible \\
        &  \cite{ Arun2021Systematic}, \cite{ Westhofen2023Criticality} & Metrics may be too simplistic, not capturing all aspects of risk \\
        &   & Choosing appropriate thresholds may be difficult \\ \hline
 
        Deriving edge cases using probability estimation &
        \cite{ Heidecker2019Towards}, \cite{Nakamura2022Defining}, \cite{ Muslim2023CutOut}, \cite{ DeGelder2023Quantitative} &
        Provides reasonably foreseeable edge cases \\
        (\cref{subsec:probability}) &   & Provides parameter ranges \\
        && Requires strong assumptions or substantial amount of data \\
        && Focussing only on parameter values \\ \hline
 
        Deriving edge cases using machine learning &
        \cite{ Laxhammar2014Online}, \cite{Pang2021Deep}, \cite{ sun2021corner}, \cite{ Ryan2021EndtoEnd}, &
        Learning from data (labelled and unlabelled)\\
        (\cref{subsec:machinelearning}) &  \cite{ Osman2019Prediction}, \cite{ Sonntag2024Detecting}, \cite{ moradloo2024safety} & Can handle high-dimensional data \\
        && An anomaly is not necessarily an edge case \\
        && Lack of clarity in why something is labelled as an edge case \\ \hline
 
        Challenging the system under test &
        \cite{ Wang2018Traffic}, \cite{DeGelder2017Assessment}, \cite{ Xu2018Accelerated}, \cite{ Zhao2018Accelerated}, &
        Depends on the system \\
        (\cref{subsec:challenging}) &  \cite{ Jesenski2020Scalable}, \cite{ Li2021Theoretical}, \cite{ Abeysirigoonawardena2019Generating}, \cite{ Althoff2018Automatic}, \cite{ liu2024safetycritical} & Requires knowledge of the system (via assumptions or simulations) \\ \hline
 
        Deriving edge cases using novel scenario generation &
        \cite{ sun2021corner}, \cite{xiong2025path}, \cite{ xu2025diffscene}, \cite{ russell2025gaia}, &
        Controllable and targeted generation of rare scenarios \\
        (\cref{subsec:novelgeneration}) &  \cite{ huang2025cadre}, \cite{ ondemand2025scenario}, \cite{ peng2026ldscene} & Realism of generated scenarios is hard to guarantee \\
        && No ground truth for assessing edge case coverage \\
        && Persistent sim-to-real gap \\ \hline
    \end{tabular}
\end{table*}

\subsection{Surrogate Metrics of Safety}
\label{subsec:metrics}
Safety surrogate measures were initially developed to predict accident rates without relying on accident data \cite{Davies2011Outline, Tarko2018Estimating}. These measures show promise for edge case detection in \ac{av} due to their ability to capture critical and uncommon safety-related events \cite{Arun2021Systematic, Wang2021AReview, Westhofen2023Criticality}.  While edge cases encompass both safety-critical and non-safety-critical rare scenarios, safety surrogate measures explore the safety-relevant subset of edge cases.

\textcite{ke2017cost} showed the effectiveness of \ac{ttc} in detecting vehicle-pedestrian near misses. They found optimal results when identifying interactions with a \ac{ttc}$<$\SI{2}{\second}. It suggested \ac{ttc}'s potential as a metric for distinguishing edge cases from regular scenarios. The \ac{btn} is another established metric \cite{Brannstrom2008Situation}, which represents the ratio of required to maximum achievable deceleration to avoid a collision.  \textcite{Asljung2016Comparing} compared \ac{btn} and \ac{ttc}, concluding that \ac{btn} is preferable due to its greater stability over time compared to \ac{ttc}'s higher variability and oscillations.

Ponn et al.\ \cite{Ponn2020Identification} introduced a ``complexity metric,'' which is based on the most complex scene of a scenario. The complexity of the scene is defined based on various factors (see \cite{Yu2020Method}), such as \ac{btn}, the number of surrounding individuals, and the occluded area for the ego vehicle. They showed that scenarios with a high value for the complexity metric are more difficult for an AV to deal with. An alternative method for quantifying the complexity of a scene is proposed in \cite{Wang2018Traffic} using support vector regression. Although the verification of their complexity metric is limited to three qualitatively assessed examples, the proposed method could be used to automatically find challenging scenarios in a dataset.

While safety surrogate measures provide a quantifiable means to predict hazardous events, their reliance on predefined thresholds may not capture the full dynamics of every scenario. Continuous metric refinement combined with holistic data strategies is essential, and synergy between metrics-based and ML methods could yield more robust detection frameworks. The exploration of new metrics that can capture other dimensions of risk, such as the psychological or behavioural state of drivers, could offer deeper insights into the myriad factors influencing vehicle interactions and safety.

\subsection{Probability Estimation}
\label{subsec:probability}
Edge cases can be seen as scenarios near the ``edges'' of the probability distribution of parameter values. \textcite{Nakamura2022Defining} used this idea to determine the ``reasonably foreseeable'' range of parameter values, a phrase from the UNECE regulation on automated lane-keeping systems \cite{ece2021WP29}. The approach assumes scenario parameters are independently distributed according to the Beta distribution. From this, a parameter range capturing \SI{99}{\percent} of the distribution is calculated as a potential edge case. \textcite{Nakamura2022Defining, Muslim2023CutOut} applied this idea to cut-in and cut-out scenarios and  demonstrated its effectiveness in identifying relevant edge cases.

\textcite{DeGelder2023Quantitative} proposed two alternative methods to estimate ``reasonably foreseeable'' parameter values. Their first method employs non-parametric \ac{KDE}, which allows the \ac{pdf} to adapt to the data without assuming parameter independence. The second approach utilises extreme value theory and applies the generalised Pareto distribution to model extreme parameter values. These methods are evaluated via case studies involving scenarios from \cite{Nakamura2022Defining} and an additional scenario where the ego vehicle approaches a slower vehicle. In a different approach, \textcite{Heidecker2019Towards} proposed to model the trajectory parameters using a Gaussian mixture \ac{pdf}. They suggested applying novelty detection techniques from \cite{Gruhl2018Novelty}. However, this study lacks implementation details or validation.

A major limitation of probability estimation methods is the assumptions made when estimating the probability distributions. 
For instance, assuming independence among parameters may not hold in interactive driving environments where the behaviour of road users is interconnected. 
Accordingly, future research should focus on developing multi-dimensional probabilistic models that can capture the interdependencies between different scenario parameters. Techniques like \ac{KDE} \cite{DeGelder2023Quantitative} or copulas \cite{aas2009paircopula, nagler2016evading} could be employed to model the joint distribution of several variables and preserve the interdependencies. Another limitation is that these methods only consider the values of the parameters in a specific scenario, and other variations of that scenario are often ignored. 
Furthermore, the development of real-time analytics frameworks that continuously update and refine probability distributions based on incoming data from \ac{av} sensors would not only improve the reliability of the detections but also allow \ac{av} systems to adapt to changing conditions.

\subsection{Machine Learning}
\label{subsec:machinelearning}
Machine learning methods are particularly valuable when specific detection criteria or metrics cannot be defined in advance or when the nature of the edge case is unknown. They aim to uncover complex or unknown situations that significantly differ from the usual patterns in datasets. As such, machine learning techniques serve as a valuable complement to metrics-driven or knowledge-driven approaches.

\textcite{Pang2021Deep} provide an overview of generic machine learning approaches for anomaly detection.
The authors identify three primary applications of machine learning in anomaly detection: feature extraction, learning representations of normality, and end-to-end anomaly score learning. Feature extraction aims to reduce data dimensionality to facilitate the detection of anomalies or outliers. Learning representations of normality helps define and characterise typical behaviours or patterns within a dataset. End-to-end anomaly score learning directly gives anomaly scores as a result of the analysis. In addition to the above classification of machine learning techniques in anomaly detection, these techniques can be classified based on their type of supervision, which is unsupervised, supervised, and semi-supervised. Supervised learning relies on labelled data, unsupervised learning operates without labelled data, and semi-supervised learning combines elements of both.

In the realm of unsupervised feature extraction, \textcite{Laxhammar2014Online} presented the \ac{shnncad} approach for detecting anomalies in trajectories using the \ac{dhd} measure. Their method is transferable to automated driving in theory but is limited as it focuses on individual trajectories, overlooking object interactions.
\textcite{sun2021corner} focused on generating challenging driving scenarios by considering the interaction between the ego vehicle and one other object. They applied \ac{pca} for dimensionality reduction, followed by density-based spatial clustering of applications with noise (DBSCAN) for clustering and k-means for outlier detection. \textcite{moradloo2024safety} utilised an unsupervised clustering technique to identify edge cases from the crashes reported to the National Highway Traffic Safety Administration (NHTSA). They argued that edge cases frequently involve the crash partner engaging in illegal actions or encountering uncommon obstacles.

By utilising recent deep learning models for feature extraction, 
\textcite{Sonntag2024Detecting} employed two autoencoders to generate semantically rich scenario embedding vectors. The first autoencoder compresses the input data, considering the ego vehicle's dynamics, surrounding objects, and lane markings into a latent space. The second autoencoder uses recurrent neural networks to encode consecutive frames into a single vector. Standard outlier detection is then applied to these embeddings. \textcite{Ryan2021EndtoEnd} demonstrated an unsupervised learning of normality approach, using Gaussian Processes to model normal human driving behaviour. Their method analyses the steering angle and velocity ranges for each segment to identify anomalies; however, it does not consider the effects of surrounding dynamic objects.

In contrast to studies utilising unsupervised methods, \textcite{Osman2019Prediction} presented a supervised end-to-end approach for detecting near-crash events from observed vehicle kinematics data. The study hypothesizes that vehicle kinematics change significantly before near-crashes during a period called the ``turbulence horizon''. This hypothesis is tested using SHRP2 NDS time-series data \cite{shrp2_nds_2013} on vehicle kinematics (speed, acceleration, yaw rate, and pedal position). They find that a shorter prediction horizon is expected to yield higher accuracy, while a reasonable turbulence horizon should capture sufficient near-crash-related changes for effective prediction.

Despite their great potential, learning-based methods present several limitations. First, they rely on the availability of extensive, high-quality datasets, which is challenging and requires precautions and intensive data preprocessing. Moreover, these methods often lack transparency and interpretability. This can hinder trust and regulatory approval in safety-critical environments. Machine learning models require continuous updating and are prone to overfitting and potentially failing to generalise well to new, unseen scenarios. They also introduce probabilistic uncertainty, which may not guarantee consistent performance across varied conditions.

Several future directions are identified. The exploration of transfer, few-shot, and zero-shot learning allows models trained on data from one region or context to be applied to a new region or setting \cite{otovic2022intra, weber2021transfer}. Considering the wide range of studies on anomaly detection in other domains \cite{blazquez2021review, lim2021time}, this could significantly foster the application and development of machine learning methods in edge case detection in \ac{ad}. The integration of reinforcement learning offers learning from edge cases by enabling models to interact with virtual environments. The use of multi-modal data sources, such as visual, radar, lidar, and trajectory data, is another area with substantial potential. The development of federated learning models is beneficial too. These models enable \acp{av} to learn collectively while keeping the data decentralised, enhancing privacy and allowing the use of a broader array of data. 

\subsection{Challenging the System Under Test}
\label{subsec:challenging}
By leveraging insights and assumptions about the system under test, scenarios can be generated that push the \ac{av} to its operational limits. A common approach for this is importance sampling \cite{DeGelder2017Assessment, Wang2018Traffic, Xu2018Accelerated, Zhao2018Accelerated, Jesenski2020Scalable, Li2021Theoretical}. In this method, scenarios are created by sampling parameter values from a distribution known as the importance density. The importance density is determined based on simulation results or knowledge about scenarios that might challenge \iac{av}. In the former case, scenarios are ranked by their criticality using specific metrics (see \cref{subsec:metrics}), and the importance density is then derived from the most critical scenarios. In the latter case, prior knowledge is used to identify challenging scenarios, such as those requiring harsh braking or extreme steering angles.

Another method for generating challenging scenarios for \iac{av} is through adversarial learning techniques. For instance, \textcite{Abeysirigoonawardena2019Generating} proposed an automated approach to create adversarial scenarios that highlighted flaws in poorly engineered or under-trained aspects of \iac{av}. These scenarios were designed to increase the likelihood of collisions in simulations. The process worked by leveraging the current driving policy of the \ac{av} to create challenging scenarios, then updating the driving policy based on how the \ac{av} would perform in these simulated situations. This iterative process aimed to enhance the \ac{av}'s decision-making and overall performance in unexpected and demanding conditions. Similarly, \textcite{Althoff2018Automatic} aimed to generate safety-critical test scenarios by generating scenarios with a limited set of possible safe actions available to the \ac{av}. This method is particularly useful for creating edge cases that push the \ac{av} to its limits. \textcolor{black}{\textcite{liu2024safetycritical} extended this idea with deep reinforcement learning, where safety-critical scenarios are generated by sequentially editing traffic scenes and agent behaviours.}

In summary, importance sampling and adversarial learning techniques aim to generate scenarios that specifically test the robustness and response capabilities of \ac{ad} systems under extreme or unusual conditions. 
They are useful in identifying potential failures before they occur in real-world operations. 
Despite their effectiveness, these methods should be treated with special consideration, particularly in the scope and realism of the scenarios they generate. 
The predictive power of importance sampling and the efficacy of adversarial scenarios are heavily dependent on the underlying models' ability to accurately represent complex, dynamic driving environments. 
There is also a risk of overfitting, where \ac{av} systems might perform well on simulated adversarial scenarios but fail to generalise to new, untested conditions. 

To overcome these challenges, future research should develop more realistic simulation environments that account for a wider range of variables, including pedestrian behaviours, weather conditions, and other non-driving entities. Additionally, incorporating real-world data into the simulation process can help in developing more accurate models that better predict and replicate the nuances of human driving behaviours (see e.g., \cite{amini2022vista, cai2022survey}). Moving forward, a hybrid approach that combines rigorous, scenario-based testing with continuous learning and adaptation from real-world driving data may offer a more effective method for preparing \acp{av} to handle the unpredictabilities of road environments safely and efficiently.

{\cstart
\subsection{Novel Scenario Generation}\label{subsec:novelgeneration}
Generative approaches for trajectory-related edge cases have evolved through diverse methodologies to address safety-critical scenario generation. Path-driven probabilistic frameworks efficiently generate abnormal behaviours like sudden braking and lane violations by decoupling optimisation from tracking \cite{xiong2025path}, while deep reinforcement learning enables corner case generation by training aggressive background vehicle policies in high-dimensional interaction spaces \cite{sun2021corner}.
Diffusion models represent the latest advancement, using iterative denoising processes to transform noise into structured scenarios with superior stability and coverage compared to traditional GANs \cite{xu2025diffscene}. Recent implementations like DiffScene and GAIA-2 \cite{xu2025diffscene, russell2025gaia} showcase controllable generation conditioned on ego-vehicle dynamics, environmental factors, road semantics, and agent configurations, which enables scalable simulation of both common and rare scenarios. The controllability aspect allows targeted generation of specific edge cases rather than random sampling. \textcolor{black}{CaDRE generates diverse safety-critical scenarios from real-world trajectories, while on-demand scenario generation explicitly controls the desired risk level of generated scenarios \cite{huang2025cadre,ondemand2025scenario}. As an example, LD-Scene combines LLM-guided specification with latent diffusion to generate controllable adversarial safety-critical driving scenarios, illustrating the emerging integration of foundation models with scenario generation \cite{peng2026ldscene}.}

The primary challenges with these methods include ensuring the realism of the generated scenarios while avoiding overly aggressive behaviours that lack real-world validity; evaluation hurdles due to the lack of ground truth for assessing edge case coverage; and the persistent sim-to-real gap, which causes generated scenarios to not accurately transfer to actual driving conditions. Additionally, the controllability-realism trade-off presents an ongoing challenge where increased parameter control may compromise behavioural authenticity \cite{gao2025foundation}.}
\section{Knowledge-driven edge cases}
\label{sec:knowledge_driven}

\cref{sec:perception_related} and \cref{sec:trajectory_related} addressed edge case identification categories that are predominantly data-driven. In contrast, this section focuses on a descriptive approach, in which \textit{knowledge} from multiple sources can be used to identify \cite{Babisch2023Leveraging, Rosch2022Space} or generate \cite{Bogdoll2021Description} edge cases. 

Expert knowledge has been widely used alongside crash statistics to identify relevant pre-crash scenarios for AV evaluation. However, such approaches typically yield simplified representations of crash data, as the scenarios must be translated into a physical test environment. This translation generally centres on the conflict situation and key crash characteristics, such as participant speeds and collision points, with heuristic rules based on occurrence frequency and severity level applied to guide scenario selection. When the focus shifts to potential edge cases, however, a broader consideration of factors becomes necessary. Knowledge-driven approaches are better suited to this task, as they enable systematic exploration of rare but plausible scenarios that extend beyond safety-critical and accident situations to encompass challenging conditions that may cause performance degradation, unexpected behaviours, or functional limitations. 

For example, combining triggering conditions such as `rain $\geq$ 20 mm/h AND curved road radius $\leq$ 100 m AND speed $\geq$ 100 km/h AND third vehicle ahead decelerating $\leq$ $-3$ m/s$^2$' can generate safety-critical edge cases, while scenarios like `unusual road markings during construction with temporary signage' may challenge perception systems without necessarily being accident-prone. Similarly, expert knowledge can identify system-specific vulnerabilities like `vehicle turning right at night with parked vehicles obstructing cycling path view' or `novel vehicle types (e-scooters) in traffic' that exploit sensor or algorithmic limitations. In this case, heuristic rules for selection would still apply, but focusing, for example, on the minimum frequency of occurrence rather than the maximum frequency.

We categorise knowledge-driven edge cases into two classes: \ac{odd}-related, where the focus is on circumstances external to the vehicle; and vehicle-focused, where the edge cases are related to the vehicle's technology or system behaviour, focusing on the \ac{av} itself. In both cases, a multi-step process is followed. \textit{The first step} involves identifying the factors or conditions that may contribute to a particular level of criticality based on criticality phenomena or triggering conditions. `Criticality phenomena' refer to the factors or events in traffic situations that are associated with increased risk or danger \cite{Neurohr2021Criticality}. These critical situations may arise from data-driven observations or expert judgement. Such judgement may involve identifying crash causes \cite{Babisch2023Leveraging}, cataloguing scenarios and traffic signs \cite{Bagschik2017Ontology}, recognising functional insufficiencies with hazardous effects \cite{zendel2017analysing, Kramer2020Identification}, or drawing from other diverse sources, such as driving rules or common-sense knowledge. \textit{The second step} involves formalising a description of the situation using an ontology \cite{Bagschik2017Ontology}. This formalisation allows for reasoning about, combining, and exchanging information between various sources. \textit{The third and last step} consists of deciding if the identified scenario or event is an edge case for the system under test according to the adopted criteria, exposure, and severity, and classifying the identified situations as a certain type of edge case. 

\begin{figure}
    \centering
    \includegraphics[width=0.9\linewidth]{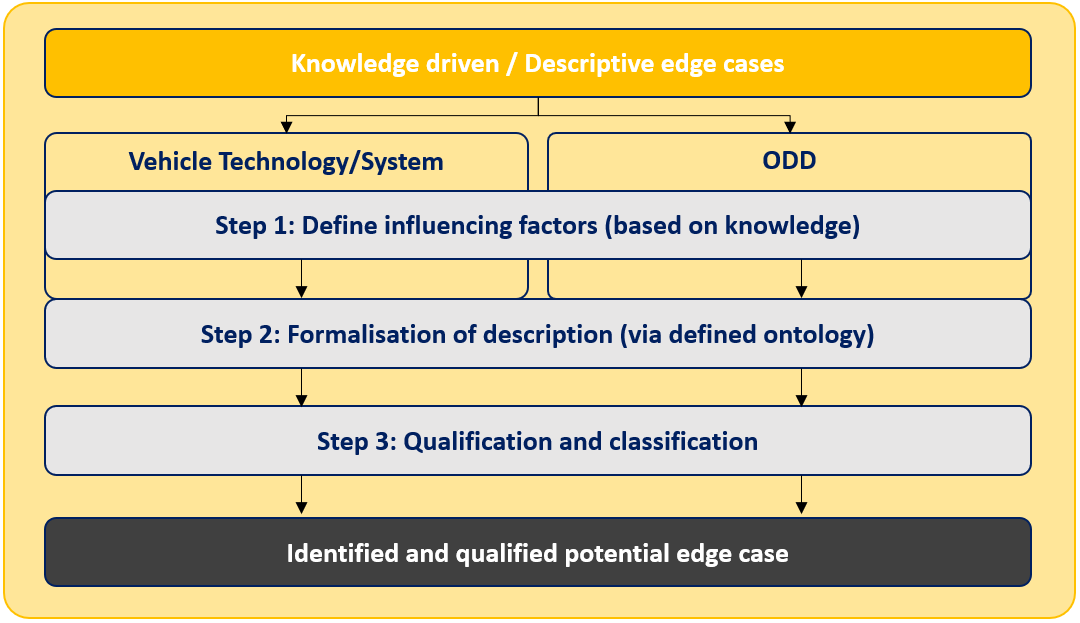}
    \caption{Steps for knowledge-based edge case identification.}
    \label{fig:Knowledge_based}
\end{figure}

These steps are depicted in \cref{fig:Knowledge_based} and discussed in the following sections. Additionally, \cref{tab:summary_knowledge_driven} summarizes the studies utilizing knowledge-driven approaches and maps them to the three steps discussed.

\begin{table*}[t]
    \centering
    \caption{Summary of the studies on knowledge-driven edge case detection}
    \small
    \begin{tabularx}{\textwidth}{@{}l>{\raggedright\arraybackslash}p{0.2\textwidth}X@{}}
    \toprule
    Step & References & Characteristics \\
    \midrule
    Step 1: Define influencing factors & 
    \cite{ Bogdoll2022AnomalySurvey}, \cite{Rosch2022Space}, \cite{Nakamura2022Defining}, \cite{Babisch2023Leveraging}, \cite{Bogdoll2021Description}, \cite{Neurohr2021Criticality}, \cite{zendel2017analysing}, \cite{Kramer2020Identification}, \cite{Bogdoll2023OneOntology}, \cite{luettin2022survey}, \cite{DeGelder2023Quantitative}
    & 
    This step aims to identify or define the influencing factors or their combination, which can lead to an edge case.
    Not all cases are edge cases in this step \\
    \midrule
    Step 2: Formalising the description via ontology &   \cite{Babisch2023Leveraging},  \cite{Bagschik2017Ontology}, \cite{Westhofen2022Using}, \cite{Bogdoll2023OneOntology} & 
    These studies discuss the need for an ontology. Most of them focus on test scenario generation (\cite{Bagschik2017Ontology}, \cite{Westhofen2022Using}, \cite{Bogdoll2023OneOntology}), while \cite{Babisch2023Leveraging} and \cite{Bogdoll2023OneOntology} discuss its relation to edge case and corner case identification as well. \\
    \midrule
    Step 3: Qualification and classification & \cite{Nakamura2022Defining}, \cite{DeGelder2023Quantitative}, \cite{Babisch2023Leveraging} & 
    This step involves validation and classification of identified potential edge cases in Steps 1 and 2. The output is the final identified edge case. \\
    \bottomrule
    \end{tabularx}
    \label{tab:summary_knowledge_driven}
\end{table*}

\subsection{Step 1: Definition of Influencing Factors}
This step deals with the definition of situations and identifying factors that can lead to a certain criticality, either based on data observations, expert judgements, or existing sources. A distinction is made between \ac{odd}-related edge cases and vehicle-focused edge cases.

\subsubsection{ODD-related Edge Cases}
Identifying \ac{odd}-related edge cases requires analysing external situations and circumstances. Possible sources include traffic rules, crash causes, the behaviour of other road users, and environmental factors. These sources can be expressed as influencing factors that describe situations potentially leading to the identification of an edge case \cite{Rosch2022Space}. Example sources are provided from the literature:

\begin{itemize}
    \item \textit{Knowledge of traffic rules} \cite{Neurohr2021Criticality}: A ``vehicle ignoring the right of way at an intersection'' is an example of not following traffic rules. Although it may be rare, it is still likely and realistic to decide that \iac{av} would have to deal with such a situation. 
    \item \textit{Crash causes} \cite{Babisch2023Leveraging}: Crashes are often multi-causal events that follow a unique sequence of events \cite{Imprialou2019Crash}, making them hard to analyse. For example, a typical crash cause might involve a combination of various factors, including environmental conditions, human error, and unexpected road conditions (e.g., sudden lane blockage). These layered causes create complex edge cases with a chain of interrelated events leading to a critical situation. 
    \item \textit{Expert knowledge} \cite{Kramer2020Identification}: This could include a common situation by adding potential hazards from its surrounding agents and environmental conditions. One example is a vehicle turning right at an intersection at night, where vehicles parked on the right side obstruct the view of a cycling path. Although the situation may seem common, further details on the situation, such as the position of the parked vehicles, may provide information about such influencing factors leading to a potential edge case.
    
    \item \textit{Synthesis through combining factors} \cite{Nakamura2022Defining, DeGelder2023Quantitative,  Bogdoll2022AnomalySurvey}: In this approach, triggering conditions or criticality phenomena can be translated into potential edge cases by combining or extending known contributing factors \cite{Bogdoll2022AnomalySurvey}. This could be done by expanding the range of parameter distributions. For example, if naturalistic driving data shows that the decelerations of leading vehicles range from \SI{-0.1}{\meter\per\second\squared} to \SI{-4.6}{\meter\per\second\squared}, this range could be deliberately extended to include more extreme decelerations, such as \SI{-6}{\meter\per\second\squared}. This can be applied to multiple parameters. 
    
    The second approach is to combine triggering conditions or criticality phenomena into more complex cases \cite{Neurohr2021Criticality, Kramer2020Identification}. An example of combining triggering conditions could be: (rain $\geq\SI{20}{\milli\meter\per\hour}$ AND curved road radius $\leq\SI{100}{\meter}$ AND speed $\geq\SI{100}{\kilo\meter\per\hour}$ AND traffic density $\geq$ 50 vehicles in view AND third vehicle ahead decelerating $\leq\SI{-3}{\meter\per\second\squared}$). In the case of the criticality phenomenon, this could be (intersecting planned trajectory AND occlusion AND non-ego violating the right of way). 
    
    A third approach is to use triggering conditions and inverse modeling from a criticality measure \cite{Bogdoll2022AnomalySurvey} to produce cases that show the behaviour of surrounding road users at an undesirable or critical level. As a simple example, searching for cases where the \ac{ttc} is less than \SI{0.6}{\second} in a cut-in scenario, the relative longitudinal speed at the start of lane change and lateral speeds are combined in such a way that a \ac{ttc} less than \SI{0.6}{\second} is produced.
\end{itemize}

\subsubsection{Vehicle-related Edge Cases}
Different from \ac{odd}-related edge cases, here the focus is on intrinsic factors and responses of the vehicle, such as wrong predictions of other road users' behaviour, or inappropriate reactions to external participants' behaviour, among others. These factors can be aligned with known system vulnerabilities to enhance the relevance of the analysis for the system under test. This approach relies on reasoning based on accident kinematics, system effect analysis, or cause-effect relationships \cite{Neurohr2021Criticality}.

The expert knowledge used in this approach can build on conventional \ac{HARA} \cite{khastgir2017towards}, by identifying possible malfunctions that could lead to system hazards and their associated risks. A systematic analysis of the physical limitations of the subsystems can be found in \cite{zendel2017analysing}. \Ac{STPA}, as a formal system engineering and hazard analysis method, can be helpful here \cite{leveson2016engineering}. The expert can use failure and incident logs of previous vehicle generations as a source. 

\subsection{Step 2: Formalisation of Edge Case Description}
The second step involves transforming the knowledge gathered earlier into a standardised or formal language. This formalisation facilitates identifying, querying, and linking the information to data sources \cite{Westhofen2022Using}. To achieve this, we use an \textit{ontology}, which is a structured representation of knowledge that can incorporate existing descriptions and classifications \cite{Bogdoll2021Description, guarino1995ontologies}. Ontologies are powerful tools for describing scenarios because they are understandable by both humans and machines \cite{Bogdoll2023OneOntology}. Ontologies can be expressed using formal languages like the Web Ontology Language (OWL) \cite{bechhofer2004parsing} or through knowledge graphs (KGs) \cite{hogan2020knowledge,luettin2022survey}. Previous research has utilised ontologies to define various types of scenarios, particularly for generating test cases \cite{Bagschik2017Ontology}, \cite{Bogdoll2023OneOntology}, \cite{Westhofen2022Using}. \textcolor{black}{More recent ontology-based work also encodes rules of the road to generate verification and validation scenarios, strengthening the link between formal scenario descriptions and executable ADS tests \cite{dacosta2024ontology}.} Additionally, ontologies are valuable for analysing corner and edge cases \cite{Bogdoll2023OneOntology, Babisch2023Leveraging}, such as situations with specific conditions (e.g., a pedestrian crossing at night during heavy rain) or rare events (e.g., a fallen tree on the road). \textcite{Bogdoll2023OneOntology} focus on using ontologies to generate corner cases, and \textcite{Babisch2023Leveraging} explore how ontologies represent critical situations to identify relevant edge cases from crash data.

To formally represent these critical situations, we extract keywords and information that can be mapped to a database schema using defined variables and labels. It's important to note that multiple related ontologies might be in use, so mapping between them might be inevitable. Some databases, like the GIDAS\footnote{\url{https://www.gidas.org/start-en.html}} \cite{otte2012injury} or CARE \cite{yannis2009cadas} crash databases, have predefined structures with variables and labels that may not match the terminology or concepts used in the ontology. To bridge this gap, the ontology must be formulated in a specific database language \cite{Babisch2023Leveraging} using tailored queries. Applying this method to monitoring data (such as field operational tests or in-service monitoring) or external data (like social media data) requires mapping events or labels from the data stream to the ontology. For monitoring data, relevant tags can be added by drivers or operators---similar to manual annotation---or through automatic labeling using a scenario detection tool. Manual annotation is particularly valuable due to the expertise of individuals directly interacting with the technology, especially when interpreting unique and novel situations.

\subsection{Step 3: Qualification and Classification}
Since critical scenarios are not always edge cases, additional criteria must be defined to qualify whether the identified cases are actual and relevant edge cases for the system under test. These criteria can include the exposure and severity of the identified situations, synthesis through combining factors, or direct expert knowledge. Examples of these criteria from the literature are provided below:
\begin{itemize}
    \item \textit{Exposure and severity of identified situations} \cite{Babisch2023Leveraging}: When verifying the edge cases for the system under test, considering both exposure (how frequently a scenario occurs) and severity (the potential impact) is important. Scenarios with low exposure but high severity are prime candidates for edge cases. Therefore, scenarios that involve a high number of critical factors, or those with unusual combinations of phenomena, are more likely to be edge cases due to their rarity or severity \cite{Babisch2023Leveraging}. Utilising risk assessment frameworks that incorporate exposure and severity could be useful to systematically identify and prioritise edge cases \cite{chia2022risk}.

    \item \textit{Defining reasonable parameter ranges} \cite{Nakamura2022Defining, DeGelder2023Quantitative}: For this criterion,  first, typical distributions of parameters (e.g., speed and distance) are captured from real-world datasets; next, the thresholds are set based on scenario frequency and probability.  Statistical extrapolation extends these ranges to cover rare but plausible scenarios, while risk acceptance criteria exclude highly unlikely events. This approach includes edge cases that are rare yet reasonable to ensure a comprehensive evaluation of system performance.

    \item \textit{Expert review}: Further expert review is needed to curate the potential edge cases to decide whether they should be adopted as true edge cases. This can be done using criteria like relevance for the system under test, sufficiently rare to be considered an edge case, or contributing additional phenomena or aspects to the current scenario collection. Over time, some cross-organisation consensus may be built up to define which cases should be considered a potential edge case.
    
    Once the criteria have been clearly defined, the edge cases can be further classified into perception-related edge cases or trajectory-related edge cases (as done in this work), or lower-level subclasses, such as unusual objects, exceeding system limits, extreme values, or anomalies. 
\end{itemize}
While significant work has been done on the individual steps (see \cref{tab:summary_knowledge_driven}), only \textcite{Babisch2023Leveraging} conceptually address all three steps while focusing on crash data analysis. In Step 1, they consider a combination of possible influencing factors, including crash causes and knowledge of traffic rules. In Step 2, they focus on translating the influencing factors into an ontology that could be applied to the used crash data. In this step, their work is limited to their used dataset since it uses a specific codebook and language. Other approaches listed for Step 2 in the table do not have such limitations and have gone further in the representation of relevant situations. In Step 3, \textcite{Babisch2023Leveraging} demonstrate the results of applying the ontology and suggest approaches for qualifying edge cases based on exposure or severity. 

{\cstart Overall, the identification of edge cases presented in this section provides a qualitative approach, where knowledge from various sources can be used in a systematic manner to identify relevant edge cases without the need for large amounts of data. However, the lack of real-world validation may introduce the risk of overlooking relevant scenarios and generating imaginary edge cases. As a mitigation, real-world data should be used to verify the existence or reasonable probability of a knowledge-driven edge case. Also, a combination with quantitative methods, such as trajectory-based approaches, would offer multiple synergistic opportunities, such as validating knowledge-driven scenarios through empirical analysis, using expert knowledge to guide trajectory-based search strategies, enriching qualitative scenarios with quantitative parameter distributions, and enhancing the coverage by capturing edge cases that neither approach would identify independently. This would require that both metadata and trajectory-based information be available. One example could be the usage of the crash dataset GIDAS and its subset GIDAS-PCM \footnote{\url{https://www.vufo.de/en/scenarios}}. Such integration would leverage the proactive identification capabilities of knowledge-driven methods while providing empirical validation through trajectory-based analysis, which is a promising direction for future research.} 

\section{Assessment Techniques}\label{sec:assessment}
Previous sections focused on identifying edge cases. However, there’s no guarantee, especially for data-driven approaches, that those detected cases truly represent realistic challenges for automated systems. Therefore, robust evaluation techniques are essential to ensure the reliability and relevance of identified edge cases. This section explores techniques to evaluate detection methods’ accuracy and performance. In particular, we review techniques and metrics for evaluating edge case detection, including detection performance (such as precision, recall, ROC-based measures), deployment practicality (e.g., overhead, delay), and domain-specific assessments (such as crash rates, expert review, exposure/severity analysis). We explore these assessment approaches per each category of detection methods discussed in \cref{sec:perception_related,sec:trajectory_related,sec:knowledge_driven}.

\begin{table*}[b]
\centering
\caption{Evaluation methods, metrics, and datasets for perception-related edge cases}
\label{tab:assess_perception}
\setlength{\tabcolsep}{3pt} 
\begin{tabularx}{\textwidth}{l >{\hsize=1.3\hsize}X >{\hsize=1.0\hsize}X >{\hsize=0.7\hsize}X}
\toprule
\textbf{Method Type} & \textbf{Name of Dataset or Simulation Tool} & \textbf{Metrics} & \textbf{Studies} \\
\midrule
\multirow{2}{*}{Benchmark} 
& \textbf{AV-Related:} Lost-and-Found, Road Anomaly, Road Obstacles, Fishyscapes, nuScenes, RadarScenes, Cityscapes, nuImages, SCAD, STU, OoDIS
& TPR, FPR95, AP, AUROC, AUPR, MIG, KL Divergence, mIoU, FPS, mAP 
& \cite{Vojir2021RoadCoupling, DiBiase2021Pixel-wiseScenes, Lis2019DetectingResynthesis, Sun2020Real-timeImages, ramakrishna2022efficient, Wang2021RadarTransformers, Kopp2023TacklingPointnet++, nekrasov2025spotting, nekrasov2025oodis} \\
\cmidrule{2-4}
& \textbf{Generic:} WildCapture, CIFAR-10/100, Places365, LSUN-Crop, SVHN, MS-COCO, PASCAL-VOC, NUS-WIDE, TinyImageNet, ShapeNet, UCSD Ped1/2, CUHK Avenue, ShanghaiTech, CIFAR-10-C 
& TPR, FPR95, AP, AUROC, AUPR, Detection Error, OOD Accuracy, ID Accuracy, AUC 
& \cite{Cultrera2023LeveragingDetection, Liu2020Energy-basedDetection, Wang2021CanKnow, Liang2018EnhancingNetworks, Hsu2020GeneralizedData, Sun2022DICE:Detection, katz2022training, Bai2023FeedDetection, Lin2021Mood:Detection, Masuda2021TowardAutoencoder, Lv2021LearningNetwork, Du2022Unknown-AwareWild} \\
\midrule
\multirow{3}{*}{Simulation} 
& CARLA simulator 
& Detection of attacks, differentiation between crashes and non-crashes 
& \cite{Cai2020Real-timeSystems} \\
\cmidrule{2-4}
& Udacity self-driving car simulator 
& TPR, FPR, F1-score, AUC-ROC, AUC-PRC 
& \cite{Stocco2020MisbehaviourSystems} \\
\cmidrule{2-4}
& CARLA simulator, ROS middleware 
& Detection consistency, trajectory smoothness, bounding box overlap, compute time 
& \cite{Balakrishnan2021PerceMon:Systems} \\
\midrule
Hybrid 
& Lost-and-Found, Road Anomaly, Road Obstacles, Fishyscapes, CARLA simulator, nuImages 
& TPR, FPR95, AP, MIG, KL Divergence, detection consistency, trajectory smoothness, bounding box overlap, compute time 
& \cite{Vojir2021RoadCoupling, ramakrishna2022efficient, Balakrishnan2021PerceMon:Systems} \\
\bottomrule
\end{tabularx}
\end{table*}

\subsection{Assessing Perception-related Edge Cases}
The approaches for assessing perception-related edge case detection methods can be classified into benchmark-based, simulation-based, and hybrid approaches. Benchmark-based methods leverage pre-existing benchmark datasets to evaluate performance against challenging scenarios and anomalies. Simulation-based methods utilize controlled environments to systematically test specific methods. Hybrid methods combine the two approaches to provide a more comprehensive evaluation. \cref{tab:assess_perception} presents an overview of studies utilising these methods and the popular datasets/simulation tools and related metrics. We delve into these approaches in the following subsections.

\subsubsection{Benchmark-based Methods}
The evaluation process in benchmark-based methods typically begins with the selection of appropriate benchmark datasets and metrics that reflect the challenging scenarios and anomalies the methods aim to detect. \cref{tab:assess_perception} provides the most popular datasets and metrics both in \ac{ad} and beyond. \textcolor{black}{Recent benchmarks further extend this evaluation landscape to 3D and instance-level anomaly understanding, including STU for LiDAR anomaly segmentation and OoDIS for anomaly instance segmentation and detection \cite{nekrasov2025spotting,nekrasov2025oodis}.}

Once the datasets are selected, the methods are tested on them to detect edge cases. The detection performance is measured using a variety of metrics. Common metrics include the True Positive Rate (TPR), measuring the proportion of actual positives correctly identified; \ac{fpr} at \SI{95}{\percent} TPR (FPR95), which evaluates the proportion of negatives incorrectly identified as positives when the true positive rate is \SI{95}{\percent}; and \ac{ap} and \ac{auc-roc}, which are used to assess the precision-recall balance and overall detection performance, respectively. Quantitative evaluations are often followed by qualitative analyses, where the results are visually inspected to understand the method's strengths and weaknesses. This involves comparing the detected cases against ground truth labels and assessing the clarity and accuracy of the detections.

\subsubsection{Simulation-based Methods}
Simulation-based evaluation methods provide a controlled environment by utilising tools like CARLA \cite{dosovitskiy2017carla} to simulate real-world driving conditions and introduce challenging scenarios, such as adverse weather conditions. Researchers often run episodes featuring both in-distribution and out-of-distribution examples to thoroughly test the performance of detection algorithms. The evaluation typically involves two categories of metrics: detection accuracy metrics (such as \ac{tpr}, \ac{fpr}, F1-score, \ac{auc-prc}, and \ac{auc-roc} \cite{Stocco2020MisbehaviourSystems, Cai2020Real-timeSystems}), and practical applicability metrics (like computation time and detection delay \cite{Cai2020Real-timeSystems}). Qualitative assessments may complement these metrics by identifying specific strengths and weaknesses in the simulated scenarios.

One limitation of simulation-based methods is the potential gap between simulated and real-world conditions (also referred to as Sim2Real gap). Even the most advanced simulators do not capture all the nuances of actual driving environments. Additionally, these methods often require substantial computational resources and expertise to set up and run. Improved simulation fidelity, through better modelling of real-world complexities and integration of real-world data \cite{amini2022vista} are promising directions. 

\subsubsection{Hybrid Methods}
Hybrid methods, which combine benchmark-based and simulation-based assessments, offer a more comprehensive approach that benefits from the realism of real-world data and the flexibility and control of simulated environments. In this regard, \textcite{Vojir2021RoadCoupling} evaluated their anomaly detection method using both real-world datasets (Lost-and-Found, Road Anomaly, Road Obstacles, and Fishyscapes) and simulation, focusing on quantitative metrics like TPR, FPR95, and AP, and qualitative visual results. \textcite{ramakrishna2022efficient} used CARLA and nuImages dataset and employed metrics like Mutual Information Gap (MIG) and Kullback-Leibler (KL) divergence to detect \ac{ood} instances and identify features causing detections. \textcite{Balakrishnan2021PerceMon:Systems} utilised CARLA and ROS, and real-world monitoring to evaluate detection consistency, trajectory smoothness, bounding box overlap, and compute time for YOLO object detection and DeepSORT object tracking algorithms. \cref{tab:assess_perception} provides a summary of these studies, their datasets, simulation environments, and metrics.

\subsection{Assessment of Trajectory-related Edge Cases}
The evaluations of trajectory-related edge cases in the literature have been particularly focused on those generated by machine learning-based approaches, which can be difficult to analyse due to their black-box nature.
These assessment strategies can be broadly categorised into three main approaches: simulation-based validation, validation using pre-labelled datasets, and human judgement. \cref{tab:assess_trajectory} provides an overview of these methods. 

\begin{table*}[t]
\centering
\caption{Evaluation metrics, datasets, and tools for trajectory-related edge cases}
\label{tab:assess_trajectory}
\begin{tabularx}{\textwidth}{lll>{\hsize=.75\hsize}X>{\hsize=.25\hsize}X}
\toprule
\textbf{Method Type} & \multicolumn{2}{l}{\textbf{Name of Dataset or Simulation Tool}} & \textbf{Metrics} & \textbf{Studies} \\
\midrule
\multirow{1}{*}{\makecell[l]{Simulation}} 
& \multicolumn{2}{l}{\makecell[l]{CARLA}} 
& \makecell[l]{Crash rate, crash-type distribution} 
& \cite{sun2021corner} \\
\midrule
\multirow{1}{*}{\makecell[l]{Pre-labelled\\Dataset}} 
& \multicolumn{2}{l}{\makecell[l]{Pre-labelled abnormal and normal trajectory dataset,\\ supervised learning on labelled dataset}} 
& \makecell[l]{Precision, recall, F1-score,\\ anomaly detection accuracy} 
& \cite{Laxhammar2014Online, Osman2019Prediction} \\
\midrule
\multirow{1}{*}{\makecell[l]{Human\\judgement}} 
& \multicolumn{2}{l}{\makecell[l]{Test track anomaly scores and visual inspection,\\ expert survey method with two-stage process}} 
& \makecell[l]{Visual inspection, anomaly score, expert\\ ratings on unusualness and relevance} 
& \cite{Ryan2021EndtoEnd, Sonntag2024Detecting} \\
\bottomrule
\end{tabularx}
\end{table*}

Utilising simulation, \textcite{sun2021corner} first reduced the dimensionality of the generated cases to ease the clustering of scenarios into normal cases and ``valuable'' cases. Then, the ``valuable'' scenarios were validated through simulations by comparing their crash rate and crash-type distributions with a baseline naturalistic driving environment to highlight the increased frequency and risk levels of the generated scenarios.

\textcite{Laxhammar2014Online} employed a pre-labelled dataset with trajectories classified as normal or abnormal to evaluate the performance of their method in identifying abnormal trajectories. Similarly, \textcite{Osman2019Prediction} used supervised learning with labelled data, reserving \SI{30}{\percent} for testing and employing standard machine learning metrics like precision and recall for validation. Although these two studies provided a quantitative assessment of edge case detection methods, they did not perform further validation beyond using standard machine learning evaluation metrics.

\textcite{Ryan2021EndtoEnd} calculated anomaly scores for test tracks and then visually inspected the drives to identify relevant abnormal events. While lacking clear validation criteria, this study brings domain knowledge and human perception into the assessment process of edge case detection methods. \textcite{Sonntag2024Detecting} employed an expert survey method to improve the reliability of subjective assessments. \num{21} experts rated the unusualness and relevance of regular scenarios and the detected edge cases in a two-stage survey process. \textcite{sun2024interpreting} proposed a visual analytics approach, specifically designed to assist domain experts in analysing and evaluating cases with unexpected behaviours. Their framework was assessed by a group of experts, demonstrating the potential benefits of such visual inspections in analysing edge cases. 

This overview indicates a lack of a comprehensive approach to assessing edge case relevance. Standard ML metrics measure detection performance, not whether the detected scenarios are actually challenging for an AV system. Evaluation protocols should therefore link detected edge cases to downstream AV behaviour, for instance through closed-loop simulation, rather than treating detection accuracy as a proxy for safety relevance. Moreover, hybrid approaches combining simulation, data, and expert knowledge is a promising and necessary direction. 

\subsection{Assessing Knowledge-driven Edge Cases}
Several studies emphasise the need for specific validation approaches for knowledge-driven edge cases. 
\textcite{Bagschik2017Ontology} suggest that it is important to account for the \textit{underlying implicit logic and traffic regulations} governing the scenario's context. In edge case detection, this implies that the initial step should focus on verifying the relevance and validity of the ideas used to generate scenarios. For example, creating a scenario where an intersection has the environmental properties of a motorway (e.g., speed limits and road dimensions) could result in unrealistic and misleading edge cases. Beyond verifying scenario validity, a comprehensive understanding of the driving environment is equally critical. \textcite{Westhofen2022Using} demonstrate this using drone footage to complement the vehicle's perspective, arguing that ego-vehicle data alone is insufficient to capture the full context of criticality phenomena. \textcite{Babisch2023Leveraging} extend this view and emphasise that identified criticality phenomena must be validated against diverse traffic datasets to confirm their correspondence to genuinely critical situations.

Validating knowledge-driven edge cases for realism, system relevance, and safety relevance remains a significant challenge, largely due to the dependence on expert judgement and context-specific knowledge inherent to the identification process. While criticality phenomena offer a more objective criterion, selecting which phenomena to target introduces further difficulties, including the need for highly detailed data sources. A promising direction for future research is the estimation of criticality exposure from real-world data, which would provide empirical grounding for rare event frequencies and help ensure that identified edge cases reflect actual driving conditions.

\subsection{Assessing Edge Case Relevance and Significance}
Beyond category-specific evaluations, assessing the overall relevance of identified edge cases is essential for prioritising them in AV testing and development. Drawing from automotive risk assessment frameworks such as ISO 26262, relevance metrics should integrate multiple dimensions, including exposure, severity, and criticality. Severity evaluates the potential consequences if an edge case is not handled successfully; exposure represents how often the scenario is encountered during operational lifetime; and criticality captures the challenge of preventing failures and should incorporate system recovery capability (whether the AV can safely detect, handle, or exit the scenario), mitigation complexity (resources and interventions required), and temporal urgency (time available for reaction before potential harm occurs). While exposure/frequency helps identify rare scenarios, assessing criticality ensures that high-risk edge cases requiring complex mitigation (e.g., sensor failure during emergency manoeuvres, novel obstacle types) are distinguished from lower-risk scenarios that existing safety mechanisms can handle (e.g., moderate fog with robust sensor fusion). Risk-based prioritisation frameworks adapted from automotive safety standards enable systematic evaluation of these dimensions, ensuring that development and testing resources are allocated to edge cases that pose the greatest combined challenge rather than simply the most unusual scenarios. \textcolor{black}{Modular evaluation frameworks operationalise this idea by combining multiple criticality indicators to rank and compare traffic scenarios for ADS assessment \cite{luttner2025assessing}.}

To operationalise prioritisation, edge cases are grouped into three action-oriented classes. Safety-critical cases, where hazardous outcomes are likely without mitigation, must be detected online and systematically targeted in offline test suites. Safety-relevant cases represent near-miss precursors identifiable via surrogate safety measures and merit high-priority detection, logging, and triage. Non-safety cases involve degradations in comfort, performance, or service availability that warrant monitoring to prevent service quality decline. Class assignment maps directly onto the Exposure-Criticality framework above: criticality governs tier severity, while exposure informs detection priority and logging frequency, together providing concrete operational triggers for evaluation workflows.
\section{Discussions and Future Directions}\label{sec:discussion}
This section delves into an examination of the findings presented in this survey, including the general challenges and limitations in the detection and handling of edge cases in \ac{ad}, and an outlook on future directions and recommendations for advancing research and development in this critical area.

\subsection{Challenges}
The challenges discussed in this section are presented without explicit prioritisation, as their relative importance varies with application context, deployment stage, and organisational priorities. Moreover, they are deeply interdependent: poor data quality compounds validation difficulties, while computational constraints may also limit addressing the sim-to-real gap. A universal hierarchy is therefore neither practical nor meaningful.

\subsubsection{Data Availability and Quality}
Effective edge case detection heavily relies on the availability of diverse and comprehensive datasets. However, many existing datasets are biased towards common driving conditions and lack the rare and extreme scenarios leading to edge cases \cite{rahmani2024automated}. Furthermore, data quality is a major concern. The presence of noise, inaccuracies, and incomplete data can adversely affect the performance of detection algorithms. Dataset biases also present persistent challenges. Geographic and cultural biases are prevalent, as many datasets are predominantly collected in European and North American contexts, reflecting specific driving norms, traffic patterns, road infrastructure, and regulatory environments, limiting generalisability to other regions. Temporal biases are equally significant, as datasets often fail to capture the evolving nature of road environments and emerging road users. The rapid adoption of delivery robots and novel mobility devices, for instance, creates new edge case categories increasingly common in urban environments yet absent from historical data. There is a pressing need for standardised collection protocols and high-quality datasets that capture the full spectrum of real-world driving conditions across diverse geographic regions and account for the temporal evolution of traffic environments, including rare and extreme events.

\subsubsection{Validation and Interpretation}
Despite the variety of assessment techniques discussed, the field lacks standardised evaluation guidelines and systematic validation pipelines. The use of varied metrics and criteria across studies hinders comparing detection methods, and existing metrics are typically borrowed from general anomaly detection or safety assessment domains rather than tailored to edge case relevance. The absence of unified criteria that jointly consider exposure frequency and criticality levels leads to inconsistent prioritisation. Compounding this, many detection methods, particularly those based on machine learning, operate as black boxes, obscuring the rationale behind identifications and complicating result interpretation. There is therefore a need for standardised evaluation pipelines and unified assessment frameworks that integrate exposure-criticality metrics to enable consistent and meaningful comparison of edge case detection methods.

\subsubsection{Sim2Real Gap}
While simulation offers a controlled setting for detecting and generating edge cases, it often fails to capture the full complexity of the real world \cite{hu2023simulation}. Sensor modelling remains a key limitation: simplified noise models inadequately represent hardware characteristics, such as range-dependent degradation in lidar or realistic clutter patterns in radar, and reproducing adverse weather conditions poses additional fidelity challenges \cite{li2024choose}. Human behaviour models present a similar gap, as deterministic frameworks struggle to capture the irrational and unpredictable actions commonly observed in real traffic. Consequently, edge cases identified in simulation may not accurately reflect real-world challenges, risking both false positives, where non-critical situations are flagged, and false negatives, where actual edge cases are missed.

\subsubsection{Computational Complexity}
Many advanced online edge case detection methods, particularly those based on deep learning and generative models, are computationally intensive. The high computational demands limit the feasibility of deploying these methods on the resource-constrained hardware typically used in \acp{av}. Moreover, the complexity of the algorithms can lead to increased latency, which is detrimental to the real-time decision-making required for safe \ac{ad}. Developing efficient algorithms that balance accuracy and computational requirements is essential for practical implementation.

\subsubsection{Industry Adoption Barriers} 
Beyond technical challenges, significant industrial adoption barriers exist. Data collection and labelling costs disproportionately burden smaller companies; certification requirements complicate ML-based methods given frameworks designed for deterministic systems; and cross-organisational data sharing conflicts with competitive interests. These factors often favour simpler, interpretable approaches over technically superior but harder-to-certify solutions.

The true complexity, however, emerges from the interconnected nature of these challenges. Data quality issues exacerbate sim-to-real gaps, as poor real-world datasets lead to inaccurate simulation models. Computational limits restrict the on-device processing power, which makes federated learning and large-scale model aggregation across vehicle fleets difficult. Additionally, validation difficulties are exacerbated by multiple interdependent factors: limited high-quality datasets reduce benchmark reliability, sim2real gaps undermine simulation-based validation approaches, and lack of interpretability standards compounds safety validation challenges. Overcoming these issues requires holistic strategies that consider these interdependencies rather than treating them in isolation.

\subsection{Future Directions}
The challenges outlined above point to several promising directions for future research in edge case detection.

\subsubsection{Improving Simulation Fidelity}
An essential direction is addressing the fundamental limitations of current simulation tools and reducing the sim2real gap. Sensor realism modelling must advance beyond simplified noise models to incorporate physics-based approaches that accurately replicate real hardware characteristics. For lidar systems, this includes modelling range-dependent degradation, multi-path reflections, and weather-induced artefacts, while radar simulation requires realistic clutter patterns, ghost target generation mechanisms, and interference effects. Furthermore, systematic integration of hardware-in-the-loop validation allows real sensor data injection into simulated scenarios.

Human behaviour modelling presents another critical challenge, as current agent models predominantly capture follower-leader dynamics. \textcolor{black}{ Incorporating techniques from artificial intelligence, reinforcement learning, and optimal control theory \cite{rahmani2026artificial, rahmani2023bi} can better capture human driving variability beyond simple follower-leader dynamics}. Moreover, these models can advance through probabilistic frameworks that capture irrational and unpredictable behaviours, including pedestrian jaywalking patterns, cyclist lane-changing decisions under stress, and driver reactions during emergency scenarios that deviate from optimal responses. Environmental dynamics modelling must address real-world complexities, including dynamic lighting conditions, weather transition effects, and multi-agent complex interactions. To bridge these gaps, several technical approaches show promise: world models \cite{ha2018world} can learn compact representations of environment dynamics, and generative methods and foundation models enable the creation of diverse, realistic scenarios with improved fidelity \cite{amini2022vista}.

\subsubsection{Few-shot and Transfer Learning}
Few-shot and transfer learning techniques offer powerful approaches to enhance edge case detection. These methods allow models trained on other driving scenarios or datasets to be efficiently adapted to new, potentially different datasets and domains. By leveraging pre-existing datasets and models, transfer learning reduces the need for extensive retraining and accelerates the development of edge case detection capabilities \cite{michau2021unsupervised}. This is particularly valuable given the scarcity of real-world edge cases. These techniques enable rapid adaptation to newly identified edge cases and facilitate cross-domain applicability, where knowledge from one operational context can be transferred to others.

\subsubsection{Federated Learning}
Federated learning offers a promising yet complex approach to enhance AV edge case detection by enabling collective learning from a decentralised vehicle network while addressing data privacy concerns \cite{chellapandi2023federated}. However, several challenges must be addressed for effective implementation. Communication and synchronisation of model updates across geographically distributed participants with varying network conditions are key difficulties. Privacy mechanisms, while necessary, can degrade detection accuracy, and cross-border data governance introduces additional regulatory and legal complexity. Edge case heterogeneity presents a more fundamental challenge, as certain rare scenarios may be highly region-specific and of limited value to other federation participants. Also, data and model aggregation is further complicated by heterogeneous data distributions and differing sensor configurations across participants. Nevertheless, collecting diverse regional edge cases remains important, since vehicles may be relocated across countries during their operational lifetime. Federated learning also remains broadly valuable for universal, common edge case categories, such as sensor failures, adverse weather, and unexpected object behaviours. Future research should explore hierarchical federated approaches that balance global knowledge sharing with local adaptation to regional driving characteristics, while addressing the associated technical and regulatory challenges. Multi-national and cross-company initiatives such as Hi-Drive\footnote{https://www.hi-drive.eu/} are examples of efforts that can facilitate this direction.

\subsubsection{Interpretability and Explainability}
Enhancing the interpretability of edge case detection methods is essential for building trust among stakeholders and regulatory bodies, particularly for safety-critical applications. For automated driving, interpretability serves specific safety functions by: (1) enabling rapid identification and correction of hazardous false negatives that could miss critical scenarios, (2) reducing unnecessary interventions from false positives that may compromise vehicle performance, and (3) facilitating systematic validation of detection logic against safety and regulatory requirements.
Safety-specific interpretability requirements are partially addressed in emerging standards such as ISO/TS 5083 \cite{ISO5083_2025}, which establishes principles for ensuring AI system safety and transparency throughout the development lifecycle. 
The interpretability measures should provide information on identifying which environmental  factors, vehicle behaviours, or sensor readings contribute to edge case classification. Techniques have been proposed for improving both interpretability \cite{molnar2020interpretable} and explainability \cite{angelov2021explainable} of data-driven methods, but these efforts should continue considering the accelerating use of AI in edge case detection. 

\subsubsection{Estimating Edge Case Exposure and Relevance}
Future research should also consider developing methods to estimate the exposure and relevance of identified edge cases to enhance the effectiveness of detection and evaluation processes. This involves creating statistical models to estimate edge case frequency in real-world scenarios \cite{de2021risk}, assessing their potential impact on AV performance and safety, and evaluating their contextual relevance to different operational design domains. By incorporating exposure and relevance estimation into edge case methodologies, researchers can more effectively prioritise scenarios, allocate resources, and optimise testing procedures.

\subsubsection{Collaborative and Holistic Evaluation Frameworks}
Proposing holistic evaluation frameworks that integrate various perspectives is essential for a comprehensive assessment of \ac{av} performance \cite{ece2021natm,dona2022recent}. Creating such a framework requires overcoming several critical challenges. A key issue is metrics misalignment, as component-level performance metrics often fail to correlate with overall system safety. A holistic framework must establish clear causal links between the performance of upstream components (like perception) and their ultimate impact on vehicle behaviour and safety outcomes. Furthermore, safety standards and testing protocols differ across regions, with bodies like the European New Car Assessment Programme (Euro NCAP), the U.S. National Highway Traffic Safety Administration (NHTSA), and China’s Ministry of Industry and Information Technology (MIIT) having distinct requirements. Developing a unified framework that reconciles these standards is a complex but essential task. Emerging international efforts (e.g., under ISO/TC 22) may mitigate this over time. Finally, a fundamental challenge is quantifying test coverage within a virtually infinite scenario space. Given the combinatorial explosion of variables in driving environments, it is impossible to test every possible situation. An effective framework must therefore include methods to strategically select the most critical test scenarios and to quantify the residual risk associated with the vast number of untested scenarios.

\subsubsection{Cybersecurity-aware Edge Case Detection}
An emerging and largely unexplored research opportunity lies at the intersection of traditional edge case detection and cybersecurity threat identification. While both domains share fundamental methodological approaches, they have evolved as largely separate research \cite{kim2021cybersecurity, girdhar2023cybersecurity, yousseef2024autonomous}. Cybersecurity attacks can generate novel edge cases through adversarial inputs, sensor manipulation, communication interference, or system compromise that may not be adequately captured by conventional operational edge case detection methods. Conversely, traditional edge case scenarios may create vulnerabilities that can be exploited by malicious actors, suggesting a bidirectional relationship between these domains. Future research should explore: (1) how cybersecurity attacks create edge cases requiring specialised detection approaches beyond traditional techniques, (2) cross-domain methodology transfer between intrusion detection systems and edge case detection frameworks, (3) development of unified detection architectures capable of simultaneously identifying operational and security-induced edge cases, and (4) investigation of how legitimate edge cases may create new attack surfaces or detection blind spots. This interdisciplinary approach could significantly enhance the robustness and security awareness of AV validation frameworks.
\acresetall  
\section{Conclusion}
This survey presents a comprehensive overview of edge case detection methods for \acp{av}, addressing a critical challenge in their development and validation. Through a systematic categorisation of detection approaches based on affected subsystems and underlying methodologies, we offer a structured analysis that illuminates the strengths and limitations of each method. Additionally, we introduce the concept of knowledge-driven edge cases, which has been largely overlooked in the existing literature, thus broadening the scope of edge case detection. The dual-level classification system proposed in this work not only facilitates targeted research and development but also promotes modular testing and validation frameworks that are crucial for the effective evaluation of \ac{av} systems. Furthermore, we map edge case detection methods to relevant evaluation techniques and metrics, providing a pathway for researchers and developers to assess the performance of \acp{av} comprehensively. By identifying current challenges and outlining future research directions, this survey supports the continuous improvement of \ac{av} safety and reliability. With these contributions, we hope this study guides the field towards more robust and dependable automated driving.

\section*{Acknowledgments}
The authors would like to thank all partners within the Hi-Drive project. This project has received funding from the European Union's Horizon 2020 research and innovation programme under grant agreement No 101006664.



 

\printbibliography



\section{Biography Section}
\vspace{-35pt}

\begin{IEEEbiography}[{\includegraphics[width=1in,height=1.25in, clip,keepaspectratio]{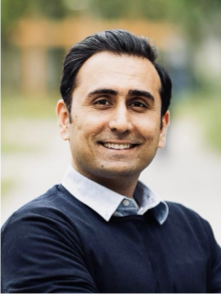}}]{Saeed Rahmani} (Graduate Student Member, IEEE) is a PhD Candidate at Delft University of Technology. During his PhD, he was an exchange scholar at the University of Oxford, a research intern at Toyota Motor Europe, and contributed to the EU Hi-Drive project. His research is focused on motion planning and decision-making for automated driving and robotics, with special interest in learning-based methods and combining them with theory-driven approaches. His work also covers traffic simulation and impact assessment of automated vehicles.
\end{IEEEbiography}
\vspace{-35pt}
\begin{IEEEbiography}[{\includegraphics[width=1in,height=1.25in,clip,keepaspectratio]{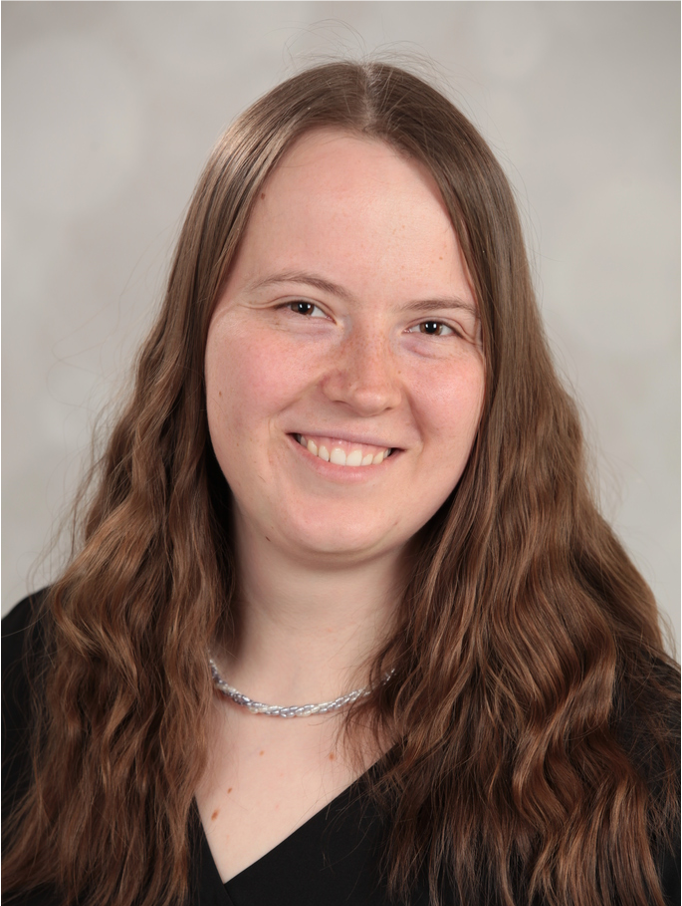}}]{Sabine Rieder} received a B.Sc. degree in informatics from the Technical University of Munich in 2019 and a M.Sc. degree from the same university in 2021. Since 2022 she has been pursuing a PhD degree at the Technical University of Munich and in 2024 she joined Masaryk University for a joint PhD program. Her research is focused on the safety of neural networks, mainly verification and runtime monitoring.
\end{IEEEbiography}
\vspace{-35pt}
\begin{IEEEbiography}[{\includegraphics[width=1in,height=1.25in,clip,keepaspectratio]{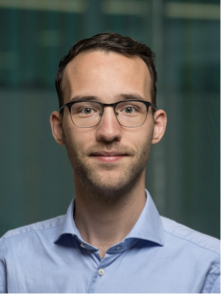}}]{Dr. Erwin de Gelder} received his MSc degree (cum laude) in Systems and Control from Delft University of Technology in 2014. Since then, he has been with the Netherlands Organization for Applied Scientific Research (TNO). His research focuses on quantitative, data-driven, scenario-based assessment methods for automated vehicles. He received his Ph.D. (2022) on this topic from TU Delft. TNO StreetWise, a tool for extracting scenarios from real-world driving data, is developed based on Erwin’s research.
\end{IEEEbiography}
\vspace{-35pt}
\begin{IEEEbiography}[{\includegraphics[width=1in,height=1.25in,clip,keepaspectratio]{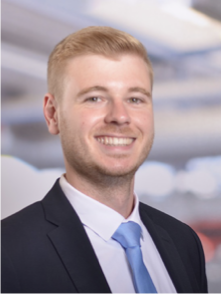}}]{Marcel Sonntag} received the M.Sc. in automotive engineering. He is currently pursuing a Ph.D. degree with the Institute for Automotive Engineering (ika), RWTH Aachen University. At ika, he is leading the research group Safety Assurance and Effectiveness. His research is focussed on detecting edge cases in trajectory data, as well as assessing the safety impacts of automated vehicles.
\end{IEEEbiography}
\vspace{-35pt}
\begin{IEEEbiography}[{\includegraphics[width=1in,height=1.25in,clip,keepaspectratio]{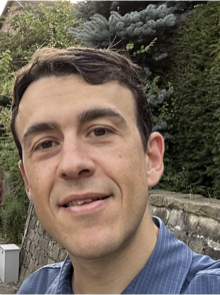}}]{Jorge Lorente Mallada} is a technical manager in the Technical Affairs Planning \& Safety Research team at Toyota Motor Europe NV/SA. He received his M.Sc degree in Industrial Engineering from the University of Zaragoza, Spain, in 2008. At Toyota Europe, he is currently in charge of safety assessment of ADAS/AD based on simulation and accident analysis.
\end{IEEEbiography}
\vspace{-35pt}
\begin{IEEEbiography}[{\includegraphics[width=1in,height=1.25in,clip,keepaspectratio]{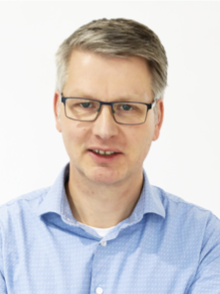}}]{Sytze Kalisvaart} is a senior project manager at TNO Integrated Vehicle Safety and initiator and product manager of the TNO StreetWise. He holds a M.Sc. TU Delft, and worked in automotive, medical, sports and recycling industry. He has a track record in usability and safety of complex systems, with a special focus on software. He has extensive experience in leading large multi partner projects, such as EU ENABLE-S3 and is coordinator of EU V4Safety on prospective safety assessment.
\end{IEEEbiography}
\vspace{-35pt}
\begin{IEEEbiography}[{\includegraphics[width=1in,height=1.25in,clip,keepaspectratio]{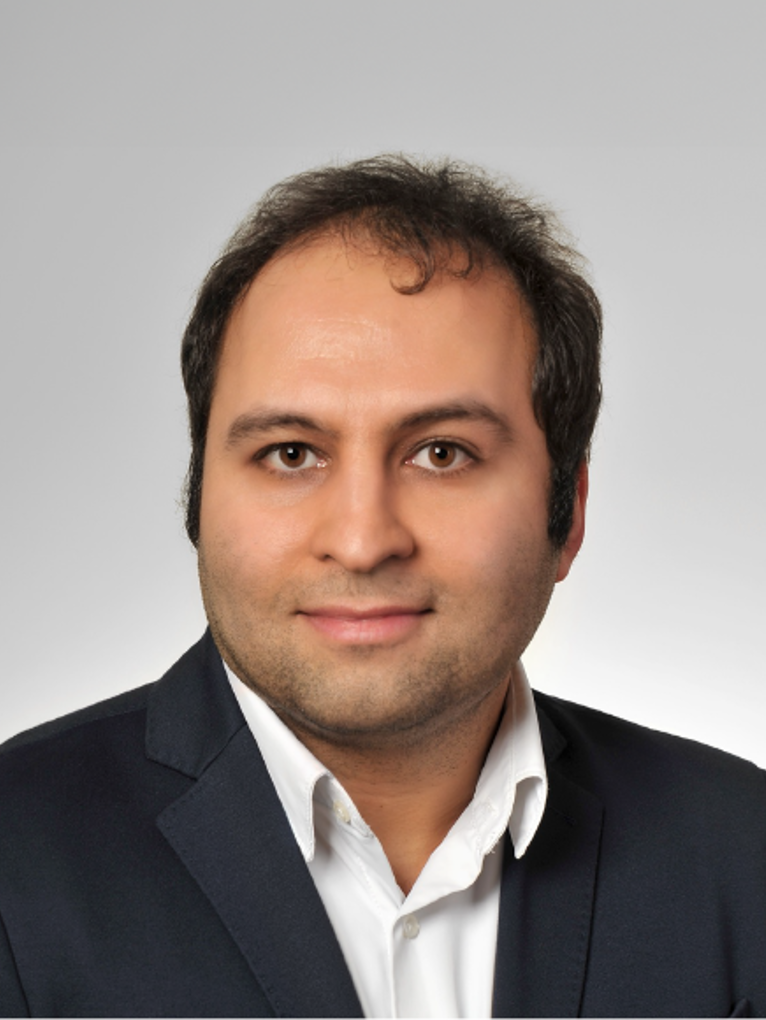}}]{Dr. Vahid Hashemi} is technical team lead at AUDI AG, responsible for developing methods and tools for safety verification and validation of learning-enabled AD systems. He's been involved as a technical lead in projects  Hi-Drive and KARLI and served as advisory board member for the EU FOCETA project. He is a committee member of the SAE G34 SG5 working group on system and safety considerations for machine learning and a member of the OECD.AI Expert Group on AI Risk \& Accountability Management.
\end{IEEEbiography}
\vspace{-35pt}
\begin{IEEEbiography}[{\includegraphics[width=1in,height=1.25in,clip,keepaspectratio]{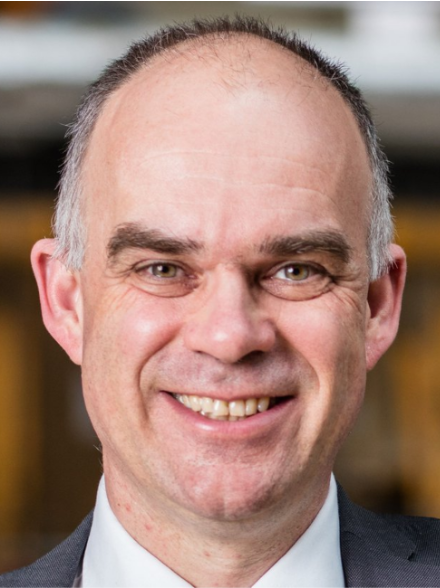}}]{Prof.Dr.ir. Bart van Arem} was appointed as full professor of Transport Modelling at the Department of Transport and Planning in 2009 and serves as Pro Vice Rector for Doctoral Affairs of TU Delft since 2021. He was head of the Department of Transport \& Planning from 2010 till 2017 and served as director of the TU Delft Transport Institute from 2012-2021. His research focuses on analysing and modelling the implications of intelligent transportation systems, such as automated, electric and shared vehicles. 
\end{IEEEbiography}
\vspace{-35pt}
\begin{IEEEbiography}[{\includegraphics[width=1in,height=1.25in,clip,keepaspectratio]{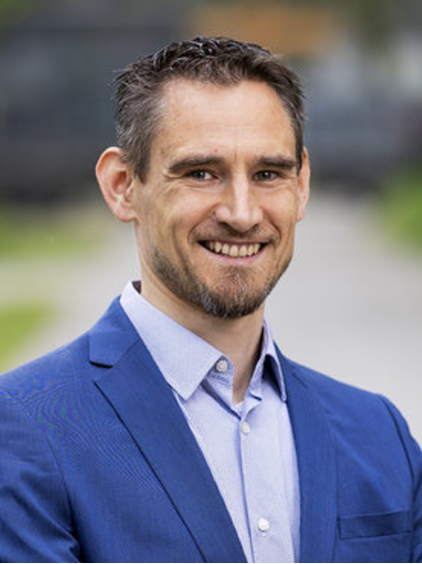}}]{Dr.ir. Simeon C. Calvert} is an Associate Professor of traffic and network management at TU Delft. He is the director of the Automated Driving \& Simulation (ADaS) Lab in the Department of Transport \& Planning and co-leads the Delft AI Lab on urban mobility behaviour: CiTy- AI. From 2010 to 2016, he worked as a Research Scientist with TNO. His research interests include ITS, impacts of vehicle automation, traffic management, traffic flow theory, and network analysis.
\end{IEEEbiography}
\vspace{5pt}

\vfill

\end{document}